\documentclass[10pt,twocolumn,letterpaper]{article}

\usepackage{iccv}              

\definecolor{iccvblue}{rgb}{0.21,0.49,0.74}
\usepackage[pagebackref,breaklinks,colorlinks,allcolors=iccvblue]{hyperref}

\def\paperID{7762} 
\def\confName{ICCV}
\def\confYear{2025}
\usepackage{graphicx} 
\usepackage{multirow}
\usepackage{booktabs}
\usepackage{threeparttable}
\usepackage{array}
\usepackage{utfsym}
\usepackage{float}
\definecolor{gray50}{gray}{0.5}
\usepackage{colortbl}
\usepackage[table]{xcolor}
\title{S$^3$E: Self-Supervised State Estimation for Radar-Inertial System}

\author{Shengpeng Wang$^1$, Yulong Xie$^1$, Qing Liao$^2$, Wei Wang$^3$\\
$^1$Huazhong University of Science and Technology,
$^2$Harbin Institute of Technology, $^3$Wuhan University\\
{\tt\small \{wsp666, yulong\_xie\}@hust.edu.cn, liaoqing@hit.edu.cn, wangw@whu.edu.cn}
}

\begin{document}
\maketitle
\begin{abstract}
Millimeter-wave radar for state estimation is gaining significant attention for its affordability and reliability in harsh conditions. Existing localization solutions typically rely on post-processed radar point clouds as landmark points. Nonetheless, the inherent sparsity of radar point clouds, ghost points from multi-path effects, and limited angle resolution in single-chirp radar severely degrade state estimation performance. To address these issues, we propose S$^3$E, a \textbf{S}elf-\textbf{S}upervised \textbf{S}tate \textbf{E}stimator that employs more richly informative radar signal spectra to bypass sparse points and fuses complementary inertial information to achieve accurate localization. S$^3$E fully explores the association between \textit{exteroceptive} radar and \textit{proprioceptive} inertial sensor to achieve complementary benefits. To deal with limited angle resolution, we introduce a novel cross-fusion technique that enhances spatial structure information by exploiting subtle rotational shift correlations across heterogeneous data. The experimental results demonstrate our method achieves robust and accurate performance without relying on localization ground truth supervision. To the best of our knowledge, this is the first attempt to achieve state estimation by fusing radar spectra and inertial data in a complementary self-supervised manner. 
\end{abstract}    
\section{Introduction}
\label{sec:intro}
State estimation techniques have garnered widespread attention as crucial facilitators in cutting-edge applications, including autonomous driving~\cite{wang2023robustloc}, indoor localization~\cite{pan2008transfer}, AR/VR~\cite{huang2024matchu,shi2023videoflow}, Robotics Navigation~\cite{hong2023learning,sandstrom2023point}, and more. Currently, optical sensors such as LiDARs~\cite{liu2023translo} and cameras~\cite{lu2021global} are considered mainstream sensors for state estimation and external perception. Numerous studies based on these sensors yield highly satisfactory outcomes under ideal conditions. However, optical sensors are not very practical in harsh conditions~\eg dust, fog, rain, snow, haze. In contrast, the millimeter-wave~(mmWave) radar, operating longer wavelengths, affords robust measurements unaffected by minuscule particles~\cite{kim2023craft}. Furthermore, mmWave radar can furnish relative radial velocity information for reflected objects, thereby indicating ego-motion and object mobility details. Leveraging these advantages, mmWave radar emerges as a promising catalyst for state estimation in diverse weather conditions. 

\begin{figure}[!t]
\centering
\includegraphics[width=.45\textwidth]{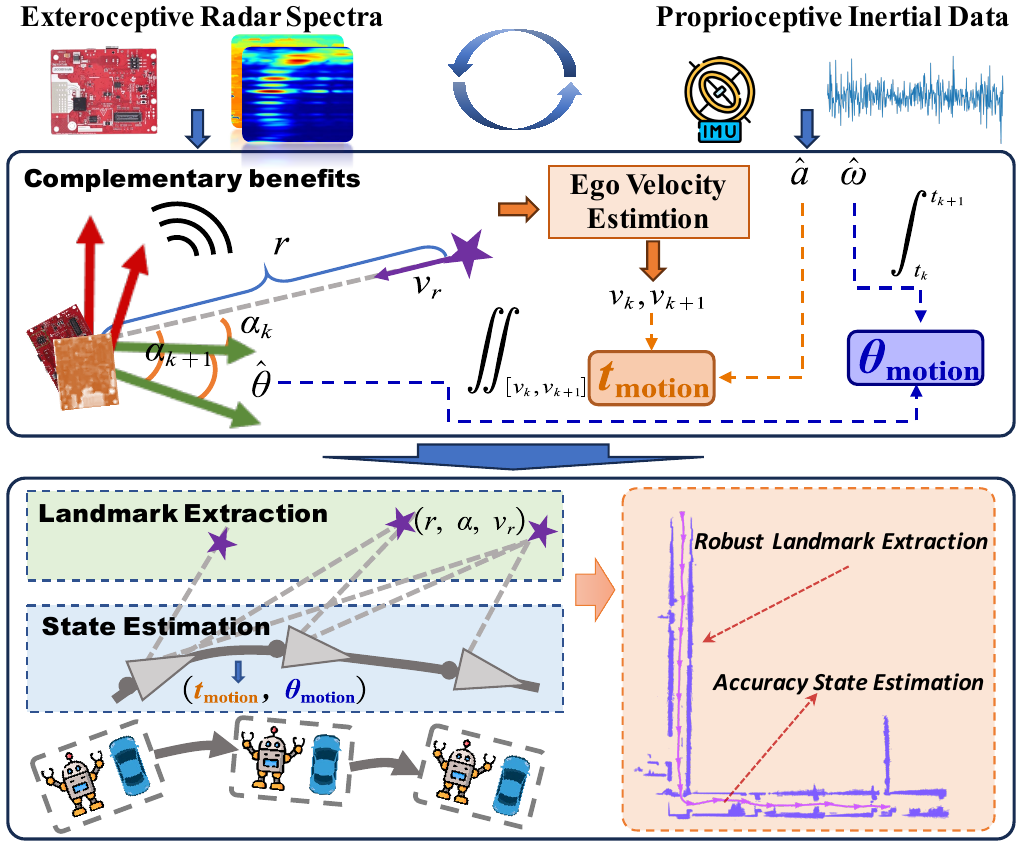}
\caption{S$^3$E fully explores complementary benefits from \textit{exteroceptive} radar and \textit{proprioceptive} inertial sensor to achieve accurate state estimation.}
\label{motivation}
\end{figure}

Existing radar-based studies~\cite{almalioglu2020milli,zhang20234dradarslam} predominantly employ the Constant False Alarm Rate (CFAR) detector to extract point clouds with Doppler velocities from the processed radar data cube, which encompasses Range-Azimuth-Doppler~(RAD) information. On this basis, some studies decouple the relative transformation between two consecutive keyframes through aligning co-observed landmarks~\cite{cen2019radar}. However, multiple objects of different sizes disrupt the independence among CFAR training cells, diminish the probability of detection, and result in missed detections~\cite{cheng2022novel}. Additionally, high-intensity cells caused by signal multi-path effects can elevate the chances of false positives, giving rise to ``ghost points". Consequently, these sparse and flawed point clouds struggle with reliable data association, posing a significant challenge for scan-matching techniques. Other studies solve ego velocity from stationary radar points with Doppler velocities~\cite{kellner2014instantaneous,doer2020ekf}. However, the ghost points and dynamic targets without spatial consistency will introduce inaccurate ego velocity estimation, degrading localization performance.

\begin{figure}[!t]
\centering
\includegraphics[width=.49\textwidth]{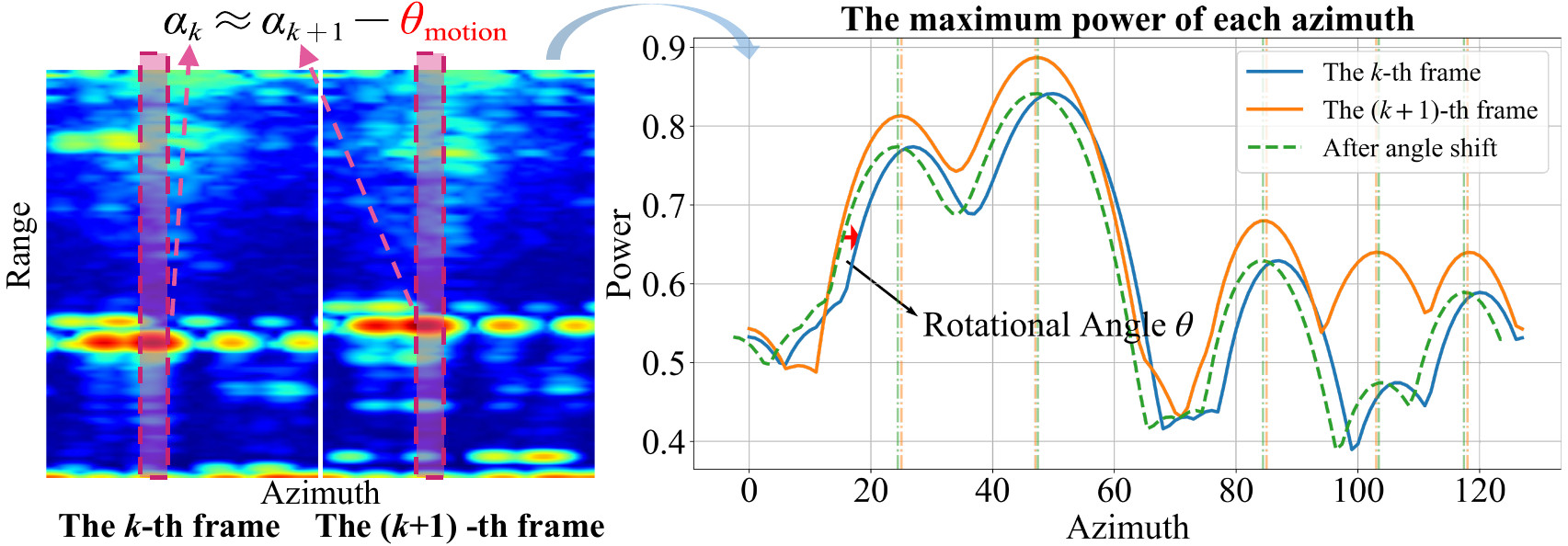}
\caption{The subtle shifts in the rotational components map into the power's linear translations in the RAS.}
\label{motivation2}
\end{figure}

Our motivation is to employ more richly informative Range-Azimuth Spectra~(RAS) to bypass sparse points, and then leverage the complementary perception capabilities of the Radar-Inertial System~(RIS) to attain robust state estimation. Specifically, as shown in Fig. \ref{motivation}, mmWave radar supplies exteroceptive information for IMU to compensate for motion cumulative drift. The proprioceptive IMU provides inertial data in kinetics to distill landmarks with motion consistency. Moreover, we observe that the quantity of principal energy moving between adjacent RAS depends on the rotational component of the motion transformation matrix. As shown in Fig. \ref{motivation2}, we take the maximum power along the azimuth for a Pow-Azimuth curve. It is worth noting that linearly translating the $k$-th curve by the motion angle yields its peak position coinciding exactly with that of the ($k+1$)-th curve. This makes rotation estimation more explicit. 

Albeit inspiring, translating this intuition into a practical and reliable state estimator is non-trivial and encounters significant issues: \textit{1) Fusion Incompatibility}. Though RAS make rotation more explicit, unlike point clouds with clear kinematic indicators, radar with inertial data cannot be directly embedded into existing fusion frameworks~(\textit{e.g.} Recursive Gaussian Filter, probabilistic factor graph optimization) due to the absence of explicit observation constraint between RAS and state factors. \textit{2) Limited angle resolution}. Commercial single-chip radars are typically equipped with a limited number of antennas, making it difficult to extract reliable landmarks from RAS.

To address the above issues, we present S$^3$E, the first state estimator that fuses radar signal spectrum and inertial data. S$^3$E fully explores the kinematic association between RAS and inertial data to achieve complementary benefits without localization ground truth. As for the limited angle resolution of radar. we propose a novel cross-fusion technique to enhance the spatial structure information by injecting inertial information into RAS. To sum up, our primary contributions are evident in the following aspects.
\begin{itemize}
    \item We propose a novel self-supervised state estimator, which leverages the complementary perception capabilities of RIS to attain accurate state estimation. To the best of our knowledge, this is the first attempt to achieve state estimation and landmark extraction from radar signal spectra and inertial data in a self-supervised manner.
    \item We propose a novel Rotation-based cross-fusion to effectively preserve motion-consistent features and enhance the spatial structure across adjacent spectra.
    \item  Experimental results demonstrate that our method outperforms the previous state of the arts and achieves complementary benefits across exteroceptive radar and proprioceptive inertial data.
\end{itemize}
\section{Related Work}
\label{sec:formatting}

Previous research on radar-based state estimation falls into two main categories: model-based and learning-based approaches. Model-based methods can be further categorized into ego-velocity-based and scan-matching-based methods. 
\subsection{Model-based State Estimation}
Many efforts follow the scan matching program from LiDAR SLAM. Cen and Newman~\cite{cen2019radar} design specific descriptors for landmarks to determine scan-matching correspondences. To mitigate outliers' effects on matching, Aldera \etal~\cite{aldera2022goes} utilize kinetic constraints to reject outliers and Burnett \etal~\cite{burnett2021we} devote to compensate for motion distortion and the measurements bias from Doppler effects. Some effects~\cite{zhuang20234d,zhang20234dradarslam} respectively operate scan-to-scan and scan-to-map as radar odometry using recently eye-catching 4D imaging radars. However, these approaches are tailored for radars with high-angle resolution and are incompatible with commercial automotive radar or mobile-integrated single-chip radar. RALL~\cite{yin2021rall} and DC-Loc~\cite{gao2022dc} propose matching schemes for aligning point clouds to pre-built maps, however, pre-built maps are unavailable in most scenarios.

Other studies focus on solving instantaneous ego velocity from single-frame point clouds of Doppler radar. Kellner \etal~\cite{kellner2013instantaneous} utilize Doppler velocities to resolve complete velocity through RANSAC~\cite{fischler1981random}. 
However, using only one radar is constrained by a single-track model under Ackermann conditions, often leading to undefined yaw rate observations. Retan \etal~\cite{retan2021radar} thoroughly analyze the explicit mapping between radial velocity and target position, assuming a constant velocity model. Additionally, they construct a radar scan cost term to optimize the pose smoothly within a sliding window. Similarly, Li \etal~\cite{li20234d} operate a 4D imaging radar with more dense measurements and derive ego-velocity pre-integration factors for pose graph optimization. While these arts achieve satisfactory state estimation performance, they often degrade or fail outright when more dynamic objects intersect the radar's scan viewsight~\cite{kellner2013instantaneous}. Moreover, landmarks without poor motion consistency retain some ghost clutter and dynamic objects, which affects velocity estimation.

\subsection{Learning-based State estimation}
Radar state estimation by deep learning is currently an emerging area of interest. Most techniques are designed for costly mechanical radars with higher spatial resolutions on par with LiDARs. However, these methods~\cite{barnes2020under,burnett_rss21,almalioglu2022deep} are incompatible with low-resolution automotive radars. In distinction, Lu \etal~\cite{lu2020milliego} achieves end-to-end supervised pose estimation. Specifically, they convert radar point clouds into a depth map, extracting spatial features by \cite{wang2017deepvo} from the depth map and IMU time series, and feeding them into the pose regression network. The paradigm achieves pose estimation in low-capacity RIS. However, due to the sparsity of radar point clouds, it falls short in extracting reliable landmarks, which can further optimize the global pose~\cite{chen2020survey}. In addition, the IMU dataflow without initial velocity cannot afford an accurate translational component of the relative transformation.
\section{Self-Supervised State Estimator for Radar-Inertial System}

\subsection{System Overview}

\begin{figure*}[htbp]
	\centering
	\includegraphics[width=\textwidth]{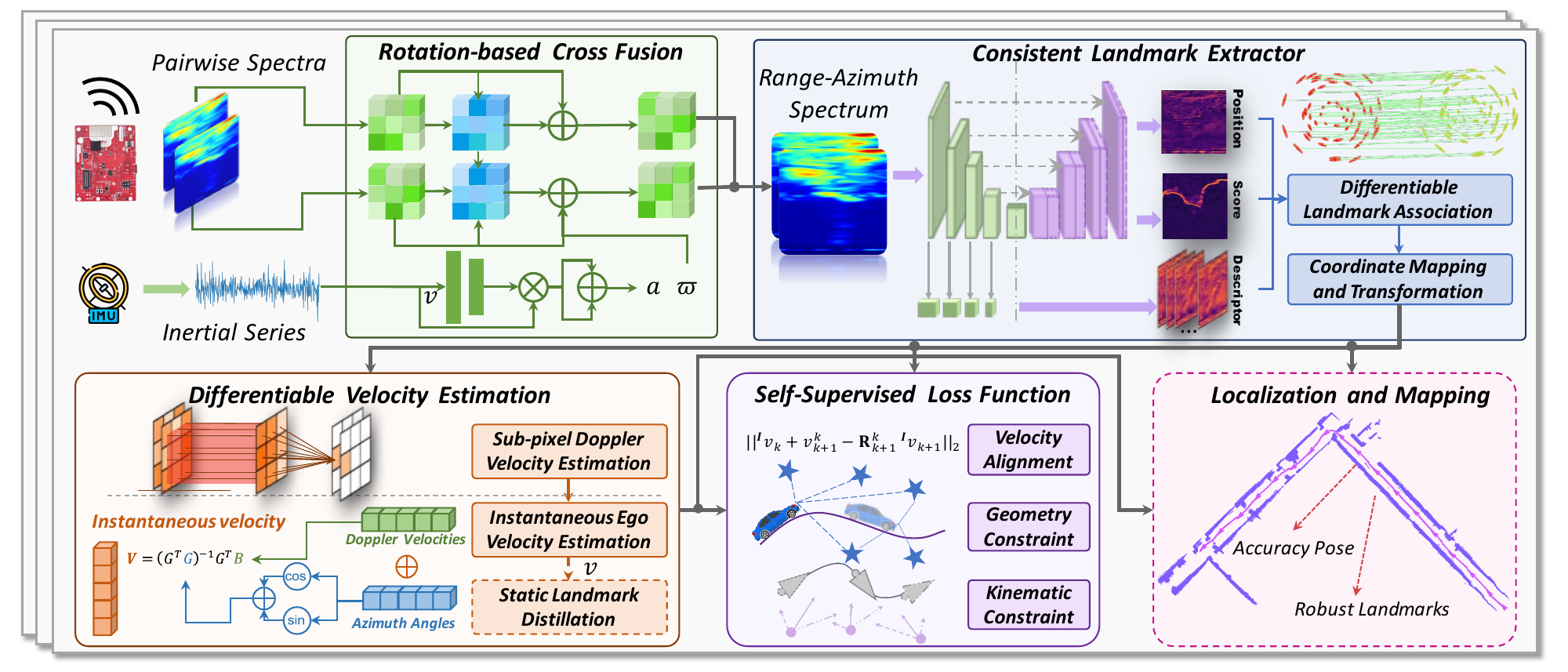}
	\caption{Overview of S$^3$E. Given a pair RAS and inertial data, it outputs accurate poses and eometry-consistent landmarks.}
	\label{sys}
\end{figure*}

In this section, the pipeline of our work S$^3$E is sequentially introduced. As shown in Fig. \ref{sys}, S$^3$E includes five modules.

\begin{itemize}
    \item \textbf{Rotation-based Cross Fusion} integrates motion information from IMU into delicately processed RAS, which preserves motion-consistent features and enhances the spatial structure information between adjacent RAS.
    \item \textbf{Consistent Landmark Extractor} predicts sub-pixel landmark locations by skip connection neural network and performs differentiable data association.
    \item \textbf{Differentiable Velocity Estimation}  estimate sub-pixel Doppler velocities for extracted geometry-consistent landmarks and solve instantaneous ego velocity.
    \item \textbf{Self-Supervised Loss Function} configures Velocity Alignment, Geometry Constraint, and Kinematic Constraint to steer the network's learning process.
    \item \textbf{Localization and Mapping} achieves robust state estimation and landmark extraction.
\end{itemize}

\subsection{Rotation-based Cross Fusion}\label{RCF}
Considering the different sampling rates of IMU and radar, IMU inter-frame pre-integration is used for data frame synchronization \cite{zhang2021conquering,qin2018vins}:
\begin{align}\label{preint}
\begin{array}{l}
	\boldsymbol{p}_{{k+1}}^{k}=\iint_{t\in \left[ k,k+1 \right]}{\mathbf{R}}_{t}^{k}\left( \mathbf{\hat{a}}_t-\boldsymbol{b}_{a} \right) \text{d}t^2\\
	\boldsymbol{\upsilon}_{{k+1}}^{k}=\int_{t\in \left[ k,k+1 \right]}{\mathbf{R}}_{t}^{k}\left( \mathbf{\hat{a}}_t-\boldsymbol{b}_{a} \right) \text{d}t\\
	\boldsymbol{q}_{{k+1}}^{k}=\int_{t\in \left[ k,k+1 \right]}{\boldsymbol{q}}_{t}^{k}\otimes \left[ \begin{matrix}
	0&		\frac{1}{2}\left( \boldsymbol{\hat{\omega}}_t-\boldsymbol{b}_{\omega} \right)\\
\end{matrix} \right] ^{\top}\text{d}t\\
\end{array}
\end{align}
where $\otimes$ is the quaternion multiplication operation, $(\cdot)^{k}_{{k+1}}$ denotes a measurement in IMU frame when receiving $(k+1)$-th RAS from mmwave radar with respect to the $k$-th IMU frame. We specify that the vehicle body coordinate system is aligned with the IMU coordinate system in this work. ${\mathbf{R}}_{t}^{k}\in SO(3)$ is an incremental rotation matrix from the current IMU frame to the $k$-th IMU frame \textit{i.e.} ${\boldsymbol{q}}_{t}^{k}$ in form of quaternion. $\mathbf{\hat{a}}$ and $\boldsymbol{\hat{\omega}}$ refer to the measurements of the vehicle body's linear acceleration and angular velocity, respectively. $\boldsymbol{b}_{a},\boldsymbol{b}_{\omega}$ are measurement biases of linear acceleration and angular velocity, which are regressed by the three-layer fully connected network with the IMU measurements and the instantaneous ego-velocity as inputs. $\boldsymbol{p}_{{k+1}}^{k}, \boldsymbol{\upsilon}_{{k+1}}^{k}, \boldsymbol{q}_{{k+1}}^{k}$ represents the increments of body's position, velocity, and orientation from the $(k+1)$-th to $k$-th frame with the $k$-th frame as the reference system. For discrete-time implementation, mid-point integration is applied.
\begin{figure}[!t]
\centering
\includegraphics[height=.85\linewidth, angle=90]{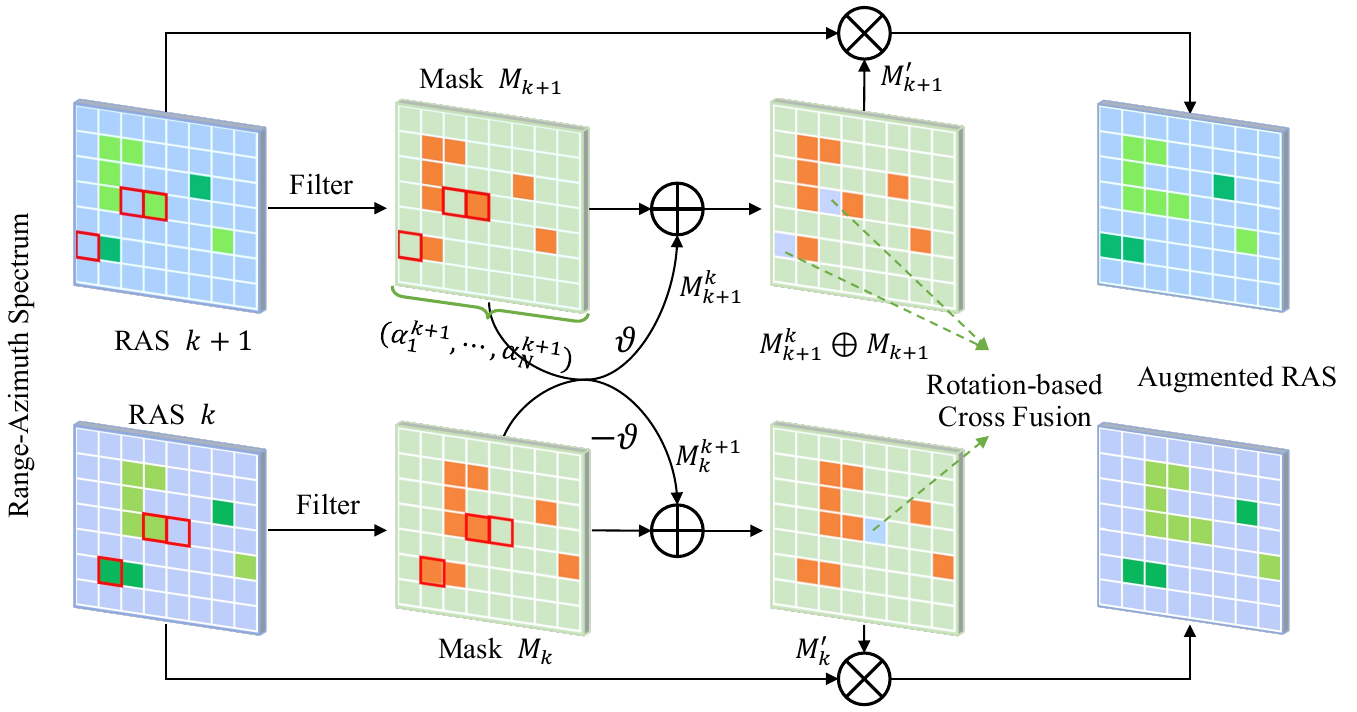}
\caption{Schematic of Rotation-based Cross Fusion.}
\label{fig_2}
\end{figure}

By pre-integration in Eqn. (\ref{preint}) aligned with radar measurement, the slight rotation variance $\vartheta$ of vehicle motion between the $k$-th and $(k+1)$-th radar frames is translated by $\boldsymbol{q}_{{k+1}}^{k}$. Next, we utilize the property that such rotation represents a linear translation between the two Range-Azimuth Spectra along the azimuth axis. The rotation-based cross-fusion between Radar and IMU is shown in Fig. \ref{fig_2}. 

Firstly, we filter noise with low power along each azimuth in RAS against speckle noise, and saturation noise picked up by the receiver. Let $\boldsymbol{M}_k= \left( \boldsymbol{\alpha }_{1}^{k},\boldsymbol{\alpha }_{2}^{k},\cdots ,\boldsymbol{\alpha }_{n}^{k} \right) \in \mathbb{R}^{H\times W}$ denote the soft mask matrix, where filtered elements are set to 0, while others are retained in the $k$-th frame RAS. $H,W$ indicates the range height and azimuth width of the RAS, respectively. Considering non-uniform angular resolution, the angle bins associated with the $n$ column vectors of the soft mask are denoted as $\boldsymbol{\eta} = (\eta_1,\cdots, \eta_n)$. One of our key insights is that the static landmarks maintain a $\theta$ orientation at the $k$-th frame and a $\theta-\vartheta$ orientation at the $(k+1)$-th frame as the radar undergoes rotation by $\vartheta$ during a trivial period. Let $\boldsymbol{\beta}^k_i$ denote the expected vector that $\boldsymbol{\alpha }^k_i$ is transformed into from the $k$-th to $(k+1)$-th frame. Expected vectors are fished by designing the dot-product attention mechanism with the angular difference as metrics:
\begin{equation}
    \boldsymbol{\beta }_m^k=\mathrm{Softmax} \left( -\frac{\left( \boldsymbol{\eta }^T-\vartheta -\eta _m \right) ^2}{\kappa} \right) \cdot \left( \boldsymbol{\alpha }_{1}^k,\boldsymbol{\alpha }_{2}^k,\cdots ,\boldsymbol{\alpha }_{n}^k \right) 
\end{equation}

where $\kappa$ indicates the temperature parameter. The expected matrix $\boldsymbol{M}_{k+1}^k=\varrho(\boldsymbol{\beta }_1^k, , \boldsymbol{\beta }_2^k,\cdots,\boldsymbol{\beta }_n^k)$ are attached to the original matrix $\boldsymbol{M}_{k+1}$ at scale $\varrho$ to enhance the landmark spatial features of the $k$-th RAS as:
\begin{equation}
\begin{aligned}
\boldsymbol{M}_{k+1}^{'}&=\boldsymbol{M}_{k+1}^k\oplus \boldsymbol{M}_{k+1} \\
\boldsymbol{M}_{k+1}^k&=\mathrm{Softmax} \left( -\frac{\left( \mathbf{1}\boldsymbol{\eta }^T-\vartheta \mathbf{1}\mathbf{1}^T-\boldsymbol{\eta }\mathbf{1}^T \right) ^2}{\kappa} \right) \cdot \boldsymbol{M}_k
\end{aligned}
\end{equation}
where $\mathbf{1}$ denotes a column vector with elements all 1. Similarly, $\boldsymbol{M}_k$ can be augmented with $\boldsymbol{M}_{k}^{'}=\boldsymbol{M}_{k}^{k+1}\oplus \boldsymbol{M}_{k}$ where $\boldsymbol{M}_{k}^{k+1}=\mathrm{Softmax} \left( -\frac{\left( \mathbf{1}\boldsymbol{\eta }^T+\vartheta \mathbf{1}\mathbf{1}^T-\boldsymbol{\eta }\mathbf{1}^T \right) ^2}{\kappa} \right)$. Such cross-fusion mapping rotation information into RAS is applied to enhance motion-consistent spatial features.

\subsection{Consistent Landmark Extractor}\label{CLE}
After Rotation-based Cross Fusion, RAS with enhanced geometry-consistent spatial features are employed to extract and match spatial feature points, serving as consistent landmarks across adjacent temporal frames. To this end, we utilize the impressive drawing power of U-Net~\cite{ronneberger2015u} for the underlying location information with deep semantic information. Specifically, we feed the enhanced RAS into U-Net as inputs and design a multi-head architecture to decode locations of feature points and acquire $N$ landmarks' locations. Assuming radar measurement uncertainty is isotropic, we decode an additional spatial weight matrix of dimensions $H \times W$ to represent scores as landmarks. Additionally, we reshape output of the encoder backbone blocks to a feature map with 248 channels before concatenating these blocks. Subsequently, we treat such feature maps with $N\times$248 dimension as descriptors of $N$ landmarks, which are used to match landmarks from different temporal frames by differentiable landmark association. Next, we specifically illustrate three headers: location, score, and descriptor.

The location header outputs detection scores $L\in \mathbb{R}^{H\times W}$ with which each pixel is selected as a candidate landmark. To fully exploit the signal energy sidelobe information in RAS, we extract sub-pixel location $(u_k,v_k)$ within a patch $\mathcal{U}_k$ centered on each pixel of $M_k$ as:
\begin{equation}
    \begin{aligned}
    &u_k=\sum_{\left( i,j \right) \in \mathcal{U}_k}{u_{ij}\left[ \text{Softmax} \left( L_{\mathcal{U}_k} \right) \right] _{ij}}\\
    &v_k=\sum_{\left( i,j \right) \in \mathcal{U}_k}{v_{ij}\left[ \text{Softmax} \left( L_{\mathcal{U}_k} \right) \right] _{ij}}
\end{aligned}
\end{equation}

The score header exports the normalized credibility weight $c_i^k$ that represents the score of each point as a landmark. This helps eliminate the effects of unpairable ghost points with highly reflective power.

The descriptor header records and concatenates encoder backbone blocks to identify the sub-pixel of the candidate landmark. Subsequently, we use the descriptors to match the candidate landmarks extracted from the previous frame into the pixel features in the current frame. Let $Q_{k-1}^N\in \mathbb{R}^{N\times 248}, K_{k}^{H\times W}$ with $(H\times W,248)$ dimension respectively denote the $N$ candidate landmarks' descriptors in the $(k-1)$-th frame and the pixel features' descriptors in the $k$-th frame. Then, the candidate landmarks' pixel location $V'$ in the $k$-th frame can be distilled through the attention mechanism \cite{vaswani2017attention}:
\begin{align}
    V'_k=\text{Softmax} \left( \frac{Q_{k-1}^{N}\cdot \left( K_{k}^{H\times W} \right) ^T}{\kappa} \right) V
\end{align}
where $V$ is a pixel location matrix of dimension $(H\times W, 2)$ by flattening all pixels $(u_i,v_j)$ in RAS. We can transform the pixel locations $V'_k$ to landmark locations $\boldsymbol{d}_{i}(i=1,\cdots, N)$ in the $k$-th radar coordinate system. With $\boldsymbol{d}_{i}$, we design a differentiable velocity estimation module to solve vehicle instantaneous velocity and further eliminate clutter and dynamic feature points.

\subsection{Differentiable Velocity Estimation}\label{DVE}
In this section, we probe the couple between the Doppler velocities of the landmarks and the vehicle's instantaneous velocity, and solve it in a differentiable way. Specifically, the proposed differentiable velocity estimation consists of the following three sub-modules.

\textit{Sub-pixel Doppler Velocity Estimation}: we use the network with concatenation skip connections in Sec. \ref{CLE} to extract sub-pixels landmark locations. To obtain the corresponding Doppler velocities, we extract the local landmark Doppler velocities from the global pixel Doppler map according to pixel coordinate distance:
\begin{align}
v^{r}=\sum_{\left( i,j \right) \in M_k}{v_{ij}^{r}\left[ \text{Softmax} \left( -\frac{\Delta _{k}^{2}}{\kappa} \right) \right] _{ij}}
\end{align}
where $v_{ij}^{r}, v^{r}, \Delta _{k}$ respectively refers to Doppler velocity corresponding to pixel $(u_i,v_j)$, sub-pixel Doppler velocities of landmarks, and differences between targets' sub-pixels and the global pixels coordinate in the $k$-th frame.

\begin{figure}[htbp] 
	\centering  
		{\label{level.sub.1}
		\includegraphics[width=0.25\textwidth]{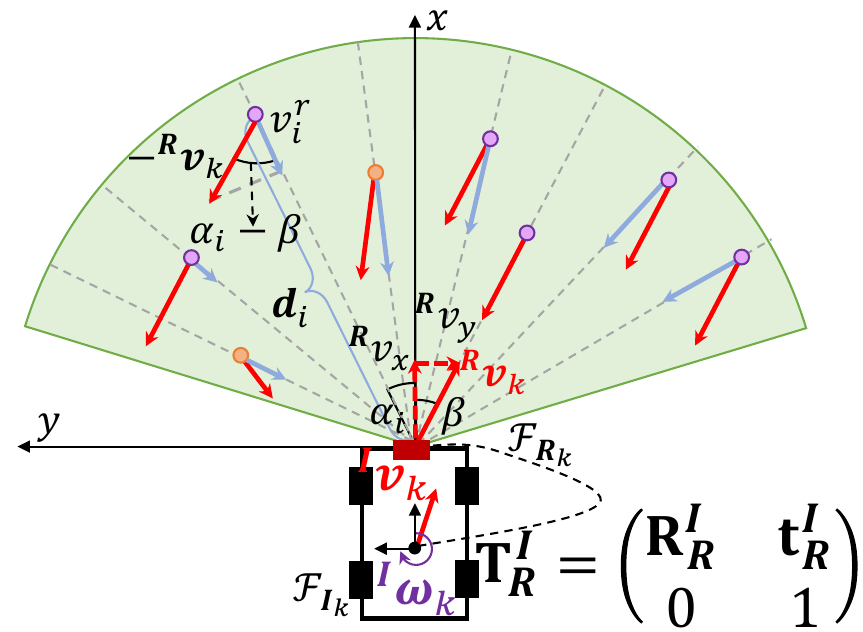}}
		{\label{level.sub.2}
		\includegraphics[width=0.21 \textwidth]{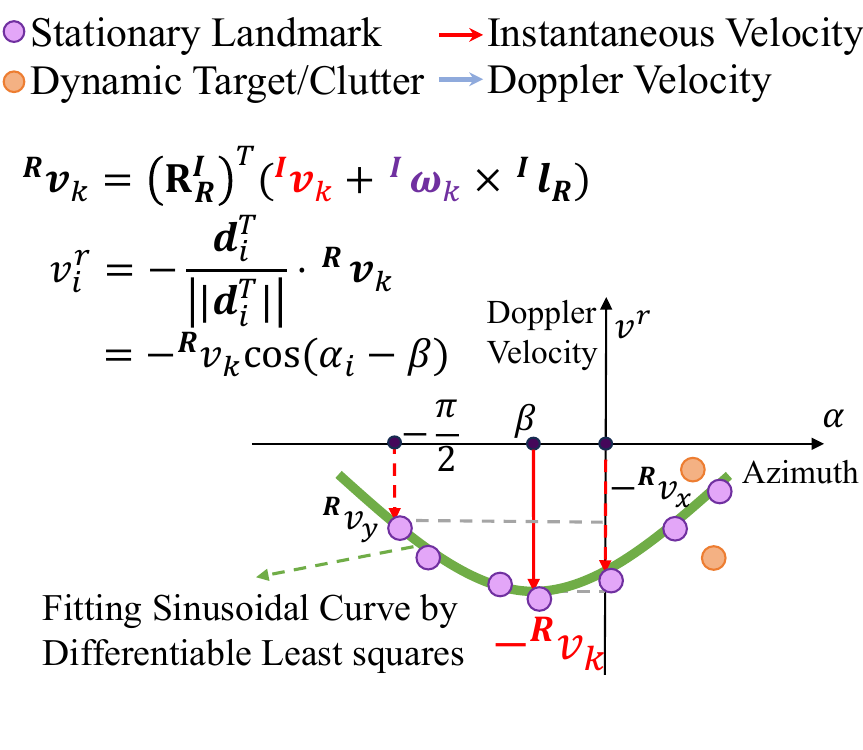}}
	\caption{Schematic diagram of estimation for the vehicle's instantaneous velocity in a differentiable manner. Left reveals a kinematic constraint between the vehicle velocity $^{\boldsymbol{I}}\boldsymbol{v}_k$, azimuth angles and relative radial velocities of stationary landmarks. The right solves the instantaneous velocity and filters dynamic targets and clutters by differentiable least squares.}
	\label{fig_velocity}
\end{figure}

\textit{Instantaneous Ego Velocity Estimation}: we set the $k$-th IMU frame $\mathcal{F}_{\boldsymbol{I}_k}$ as the body frame, which is aligned with the ego vehicle frame. The instantaneous velocity $^{\boldsymbol{R}}\boldsymbol{v}_k$ of the vehicle in the $k$-th radar frame $\mathcal{F}_{\boldsymbol{R}_k}$ can be expressed in terms of its velocity $^{\boldsymbol{I}}\boldsymbol{v}_k$ in the body frame:
\begin{align}
    ^{\boldsymbol{R}}\boldsymbol{v}_k=
\left( \mathbf{R}_{\boldsymbol{R}}^{\boldsymbol{I}} \right) ^T\left( ^{\boldsymbol{I}}\boldsymbol{v}_k+^{\boldsymbol{I}}\boldsymbol{{\omega_k}}\times ^{\boldsymbol{I}}\boldsymbol{l}_{\boldsymbol{R}} \right) 
\end{align}
where $ \mathbf{R}_{\boldsymbol{R}}^{\boldsymbol{I}}$ means the relative rotation from the radar frame to the IMU frame, $^{\boldsymbol{I}}\boldsymbol{{\omega}}_k$ refers to vehicle angular velocity, and $^{\boldsymbol{I}}\boldsymbol{l}_{\boldsymbol{R}}$ donates the radar mounting position relative to the IMU. Next, as shown in Fig. \ref{fig_velocity}, we observe a kinematic constraint between the vehicle velocity in the body frame and relative radial velocities \textit{i.e.} Doppler velocities of the stationary landmarks. We combine all cosine constraints from radial velocity observations to yield a super-definite equation about the vehicle velocity:
\begin{equation}\label{vr}
    \begin{aligned}
        -\left[ \begin{array}{c}
    	v_{1}^{r}\\
    	\vdots\\
    	v_{N}^{r}\\
    \end{array} \right] =\left[ \begin{matrix}
    	\cos \alpha _1&		\sin \alpha _1\\
    	\vdots&		\vdots\\
    	\cos \alpha _N&		\sin \alpha _N\\
    \end{matrix} \right] \left[ \begin{array}{c}
    	^{\boldsymbol{R}}\boldsymbol{v}\cos \beta\\
    	^{\boldsymbol{R}}\boldsymbol{v}\sin \beta\\
    \end{array} \right] 
    \end{aligned}
\end{equation}
where $\alpha_i, \beta$ respectively indicate the direction of landmark $i$ and ego vehicle' velocity direction in the body frame. Eqn. (\ref{vr}) can be written as $\boldsymbol{B}=\boldsymbol{G}\boldsymbol{X}$, and in consequence, the velocity can be solved in a differential way as $(\boldsymbol{G}^T\boldsymbol{G})^{-1}\boldsymbol{G}^T\boldsymbol{B}$ by least squares.

\textit{Static Landmark Criteria and Distillation}: We adopt Eqn. (\ref{vr}) as a criterion for stationary landmarks and further filter dynamic landmarks and clutter as well as mismatched pairs by RANSAC. Notably, this process is cooled down during the training phase and activated until the test phase.

\subsection{Self-Supervised Loss Function}
Constructing geometric and kinematic constraints between the proprioceptive IMU and exteroceptive radar is crucial for enabling self-supervision in RAS. A straightforward approach is to respectively decouple the inter-frame transformation matrix 
$\left[ \begin{matrix}
	\mathbf{R}_{k}^{k+1}&		\mathbf{t}_{k}^{k+1}; 
	0&		1\\
\end{matrix} \right] 
$ from the radar and the IMU as constraints. Unfortunately, the IMU inter-frame integrals are missing initial velocities, making the translational component of the transformation matrix unimpressive. In this case, we novelly propose to exploit the body velocity implicit in the radar Doppler information to compensate for the lack of initial IMU velocities. This allows us to construct constraints using the landmark points. Specifically, we design joint loss function considering the geometric constraint between IMU transformation and landmark points, the kinematic constraint applied to the Doppler measurements, and the velocity disparity between multiple sensors as follows:
\begin{itemize}
    \item \textit{Geometry Constraint}: Each landmark pair $(p_{k}^{i},q_{k+1}^{i})$, $i=1,\cdots ,N$
 from the $k$-th and $k+1$-th frames follows the IMU transformation matrix as $q_{k+1}^{i}=\mathbf{R}_{k}^{k+1}p_{k}^{i}+\mathbf{t}_{k}^{k+1}$. Hence, the geometry loss can be listed by accumulating residuals in the Mahalanobis Distance \cite{liu2021robust}:
\begin{align}
    \mathcal{L}_1=\frac{1}{2}\sum_{k\in \mathcal{B}}{\sum_{i=1}^N{c_i^ke_{i}^{T}e_i}}
\end{align}
where $\mathcal{B}$ denotes the batch number and $e_i=\mathbf{R}_{k}^{k+1}p_{k}^{i}+\mathbf{t}_{k}^{k+1}-q_{k+1}^{i} $ refers to the residual. The unimpressive translational component $\mathbf{t}_{k}^{k+1}=\int_{t\in \left[ k,k+1 \right]}{\mathbf{R}_{t}^{k}\cdot ^{\boldsymbol{I}}\boldsymbol{v}_k\text{d}t}+\boldsymbol{p}_{k+1}^{k}$ is patched by Eqn. (\ref{vr})(\ref{preint}).
\item  \textit{Kinematic Constraint}: All distilled static landmark points satisfy the cosine constraint from radial velocity observations in Eqn. (\ref{vr}), which can be described as:
\begin{align}
    \mathcal{L}_2=||\boldsymbol{G}(\boldsymbol{G}^T\boldsymbol{G})^{-1}\boldsymbol{G}^T\boldsymbol{B}-\boldsymbol{B}||_2
\end{align}
\item \textit{Velocity Alignment}: The body velocity in $(k+1)$-th frame at the $k$-th coordinate system can be expressed by combining IMU and radar observations or directly using velocity transformation. Align these expressions to get velocity loss:
\begin{align}
\mathcal{L}_3=||^{\boldsymbol{I}}\boldsymbol{v}_k+\boldsymbol{\upsilon}_{k+1}^{k}-\mathbf{R}_{k+1}^{k} \cdot \,\, ^{\boldsymbol{I}}\boldsymbol{v}_{k+1}||_2
\end{align}
\end{itemize}
Our total loss function joins the above losses in a weighted sum as:
\begin{align}
\mathcal{L}_{total}=\mathcal{L}_{1}+\lambda_1\mathcal{L}_{2}+\lambda_2 \mathcal{L}_{3}
\end{align}

Finally, in the Localization and Mapping module, Static Landmark Criteria and Distillation in Sec. \ref{DVE} is activated, and the landmark points are selected according to stationary metric in Eqn. (\ref{vr}). Subsequently, all static landmarks are aligned to the initial coordinate frame and integrated into a spatially coherent map by relative pose transformations.
\section{EXPERIMENTS AND EVALUATION}
\subsection{Experimental Setup}

\textbf{ColoRadar Dataset}.
This work focuses on extracting geometry-consistent landmark points from Range-Azimuth Spectrum, in conjunction with IMU data streams. We conduct our method on ColorRadar Dataset~\cite{kramer2022coloradar}, which contains unprocessed analog-to-digital convert sample data and provides access to RAS through signal processing. The ColoRadar contains both indoor and outdoor scenes with a total of 52 sequences and a distance of 107.4km. It furnishes LiDAR, single-chip radar (TI AWR1843BOOST-EVM), and IMU data. To validate our method's superiority, we utilize lower-resolution single-chip radar data alongside IMU data for our experiments. For a fair comparison with other learning-based baselines, we select 36 sequences as the training set and 16 for testing.

\begin{figure}[htbp]
	\centering
	\includegraphics[width=.48\textwidth]{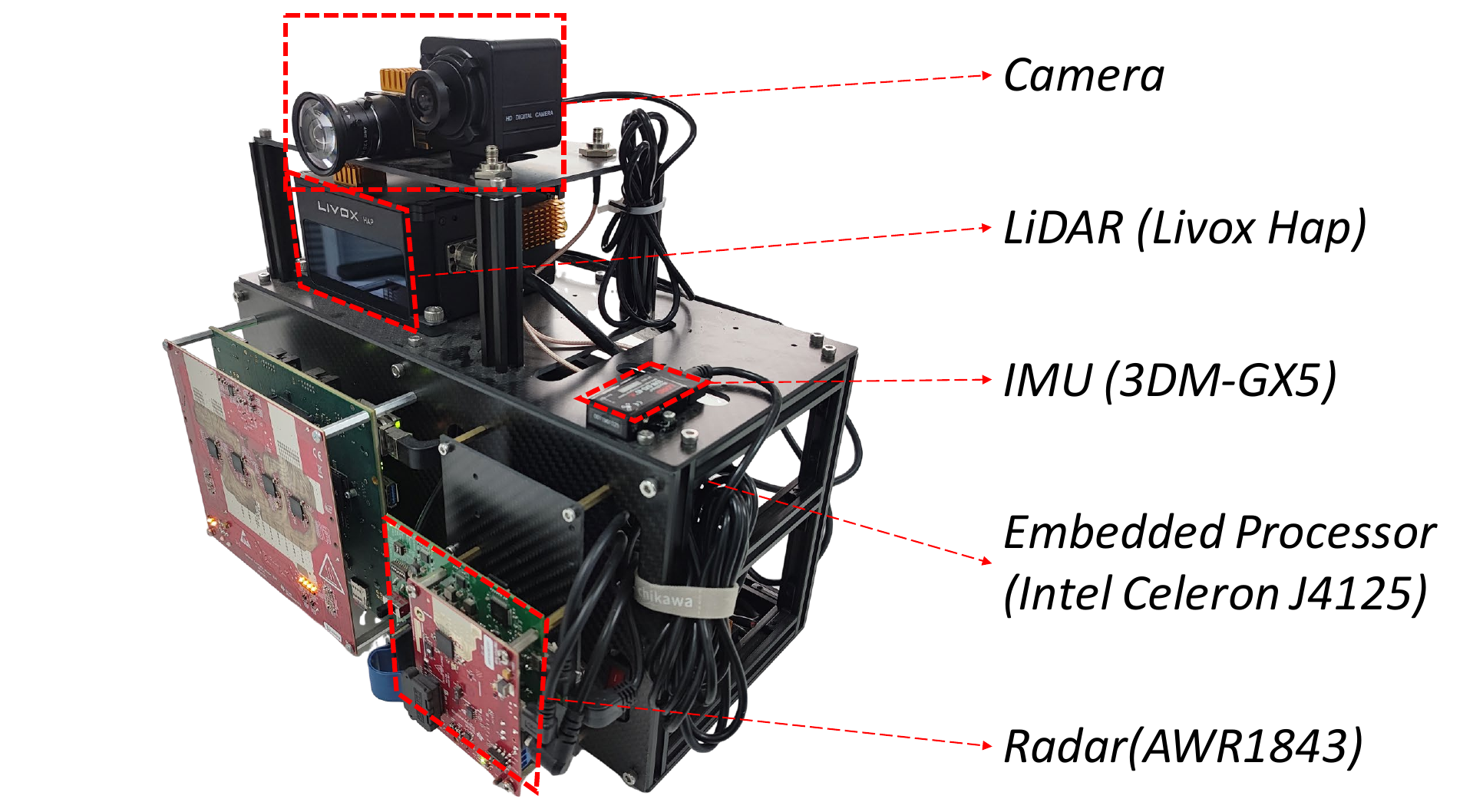}
	\caption{Our multi-modal sensing platform.}
	\label{device}
\end{figure}

\noindent \textbf{Self-collected Dataset}
We built a multimodal data acquisition platform~(Figure~\ref{device}) and collected data in 12 indoor and outdoor scenes. The localization ground truth was obtained by using Fast-Livo\footnote{https://github.com/hku-mars/FAST-LIVO}. To demonstrate the generalization of S3E as a self-supervised method over supervised learning~\cite{lu2020milliego}, we evaluate various methods in such unseen scenes.

\noindent \textbf{Implementation Details}.
We implement our system S$^3$E using Pytorch 1.11.0 with CUDA 12.4. The extractor network is fine-tuned using the pre-trained model from RadarHD~\cite{prabhakara2023radarhd} to make S$^3$E converge better. The parameters $\lambda_1$ and $\lambda_2$ of the weighted kinematic constraint and velocity alignment loss function are respectively set to 0.05 and 0.1. The temperature coefficient $\kappa$ and scale parameter $\varrho$ set to 0.01, and 0.3, respectively. We train the networks for 120 epochs on ColoRadar Dataset with Adam optimizer, which has a learning rate of $10^{-4}$ and controls it with the ReduceLROnPlateau scheduler. Specifically, The pixel values within the first six range bins of RAS are set to zero to mitigate effects of fixed installations within the radar's FOV. Moreover, to enhance spatial feature density, we interpolate each RAS from 128$\times$128 to 256$\times$256 before filtering speckle noise below 0.8 of the mean value per azimuth. It takes about 6 hours to train our model with a machine using one NVIDIA GeForce RTX 4090 and Intel Xeon Gold 6226R CPU.

\begin{figure*}[!t]
\centering
{
    \includegraphics[width=.19\linewidth]{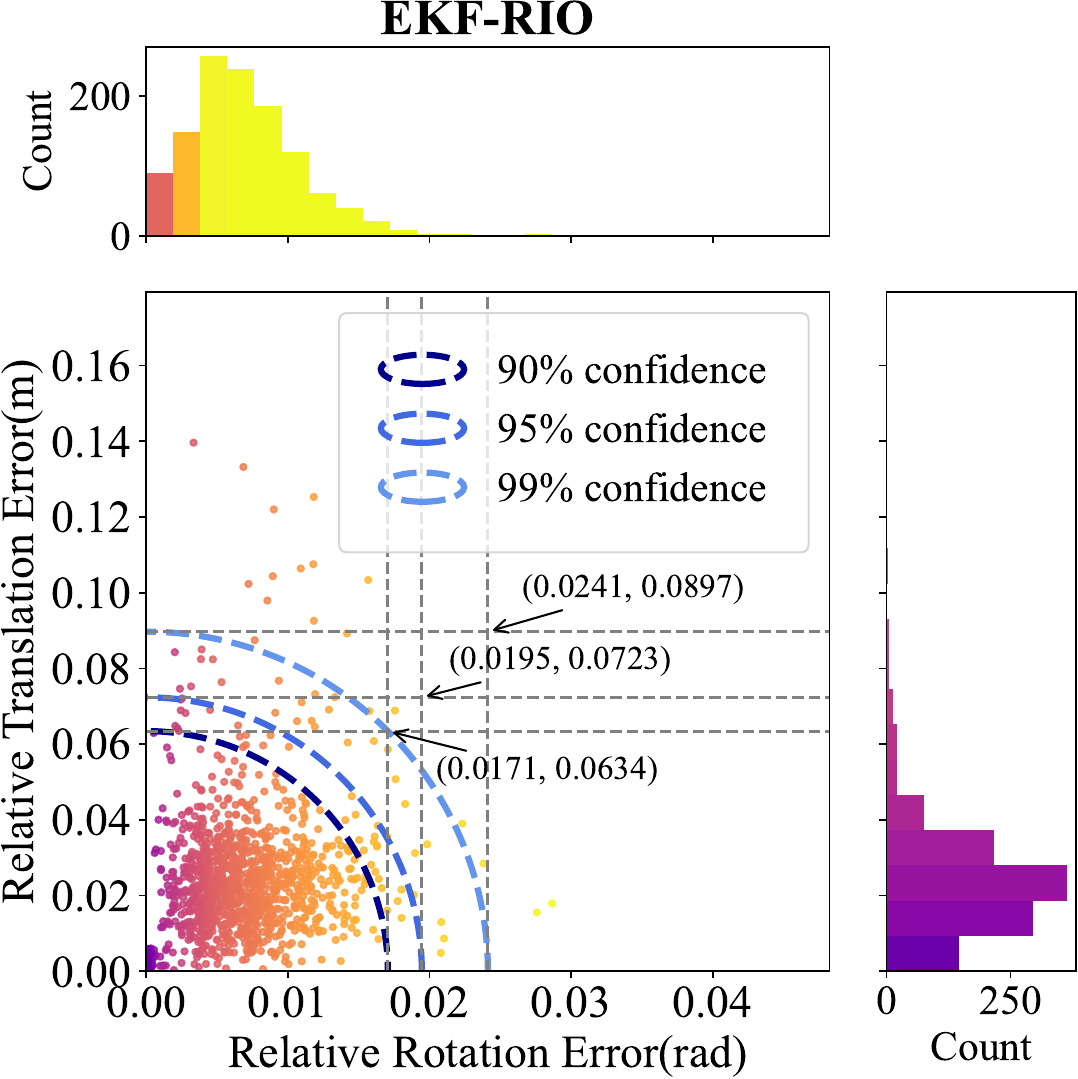}}
    \includegraphics[width=.19\linewidth]{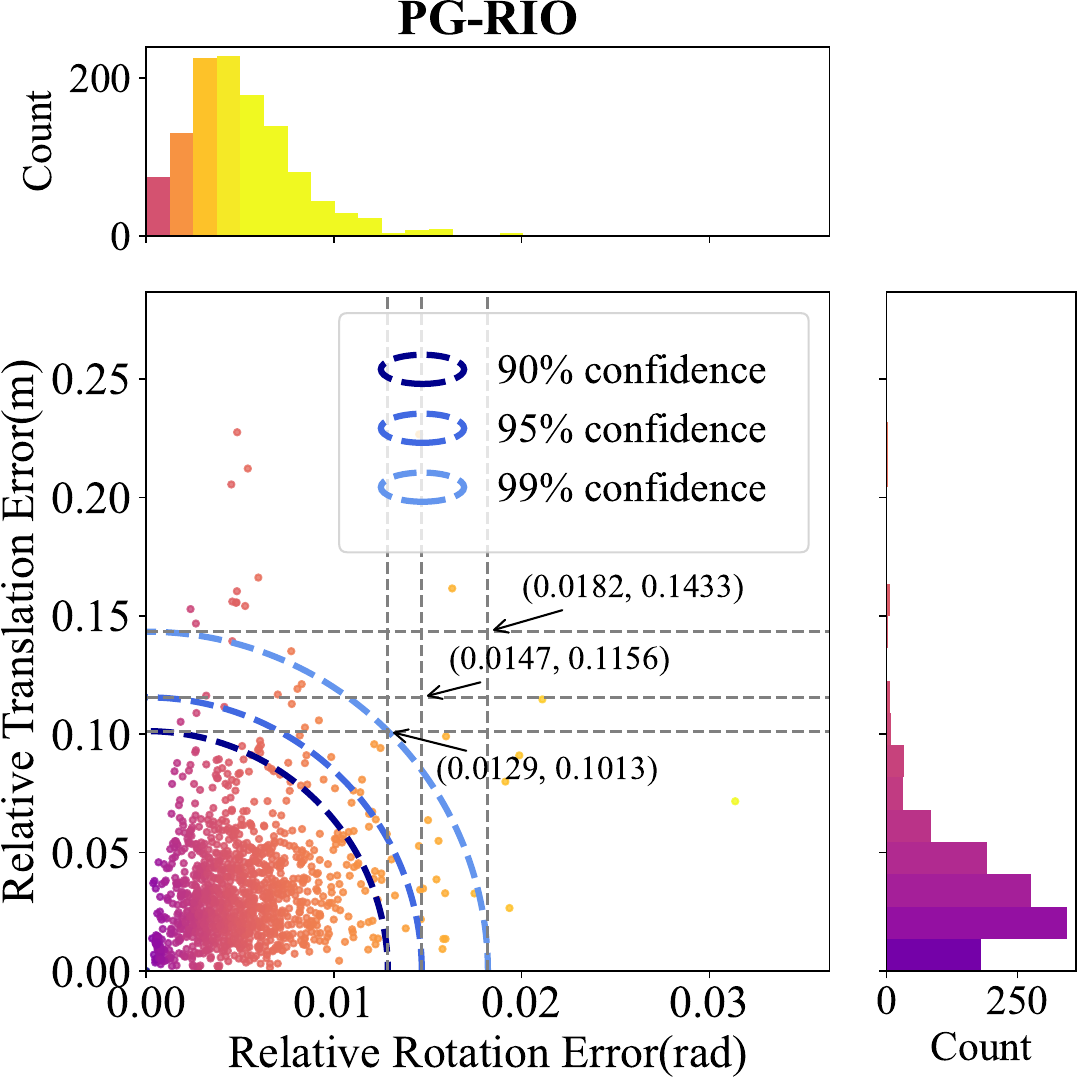}
        {\includegraphics[width=.199\linewidth]{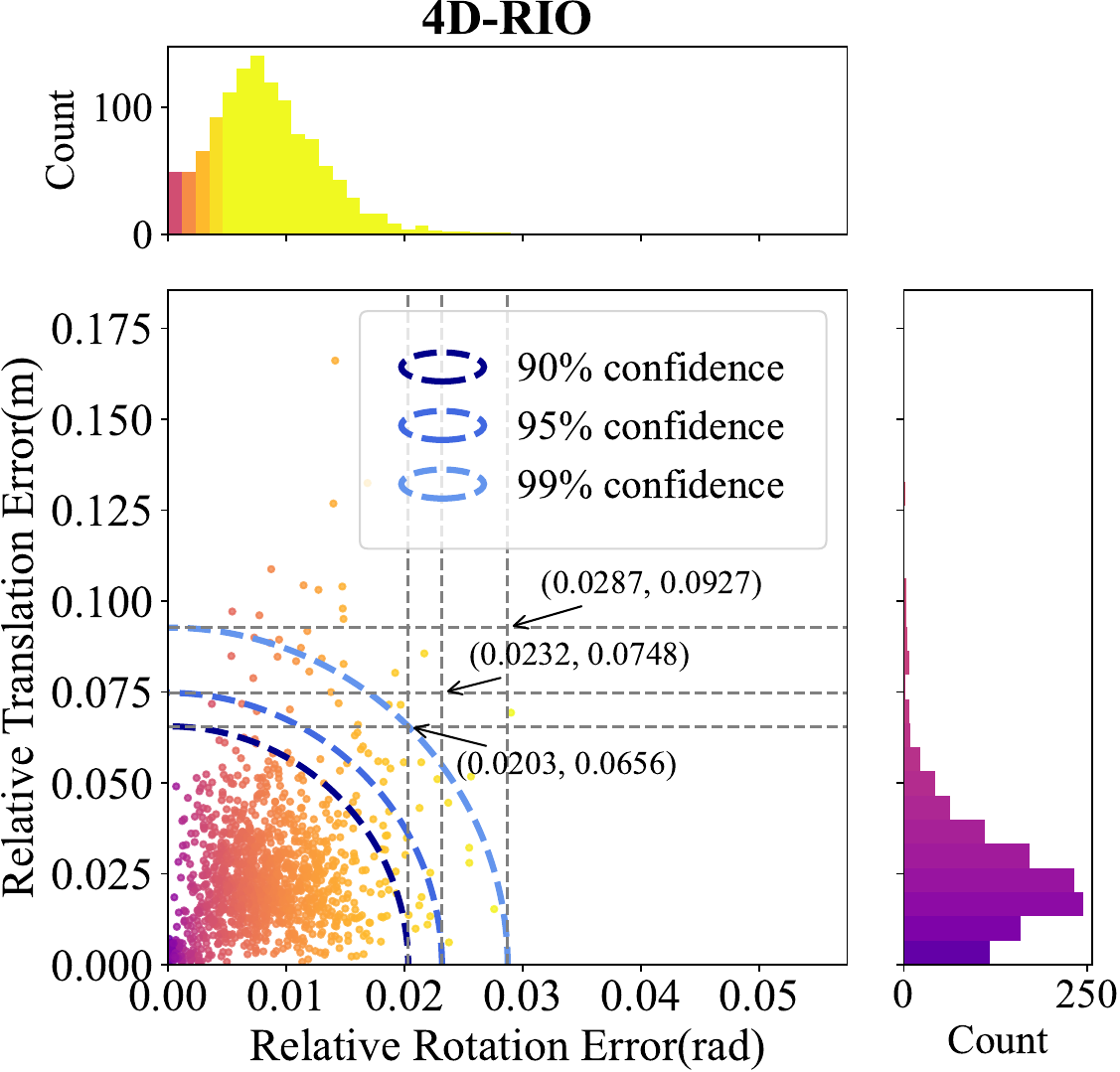}}
         {\includegraphics[width=.19\linewidth]{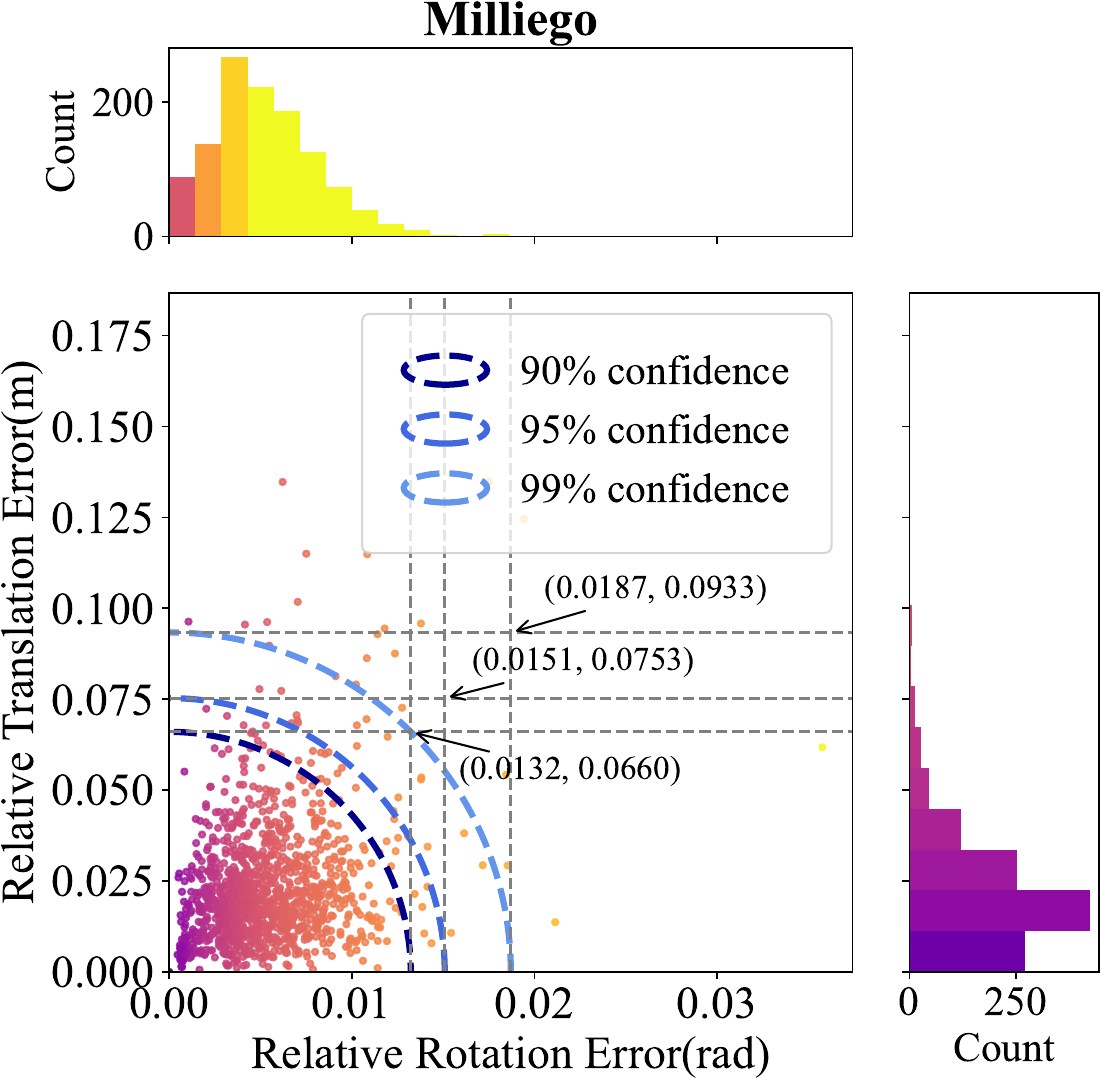}}
    {\includegraphics[width=.19\linewidth]{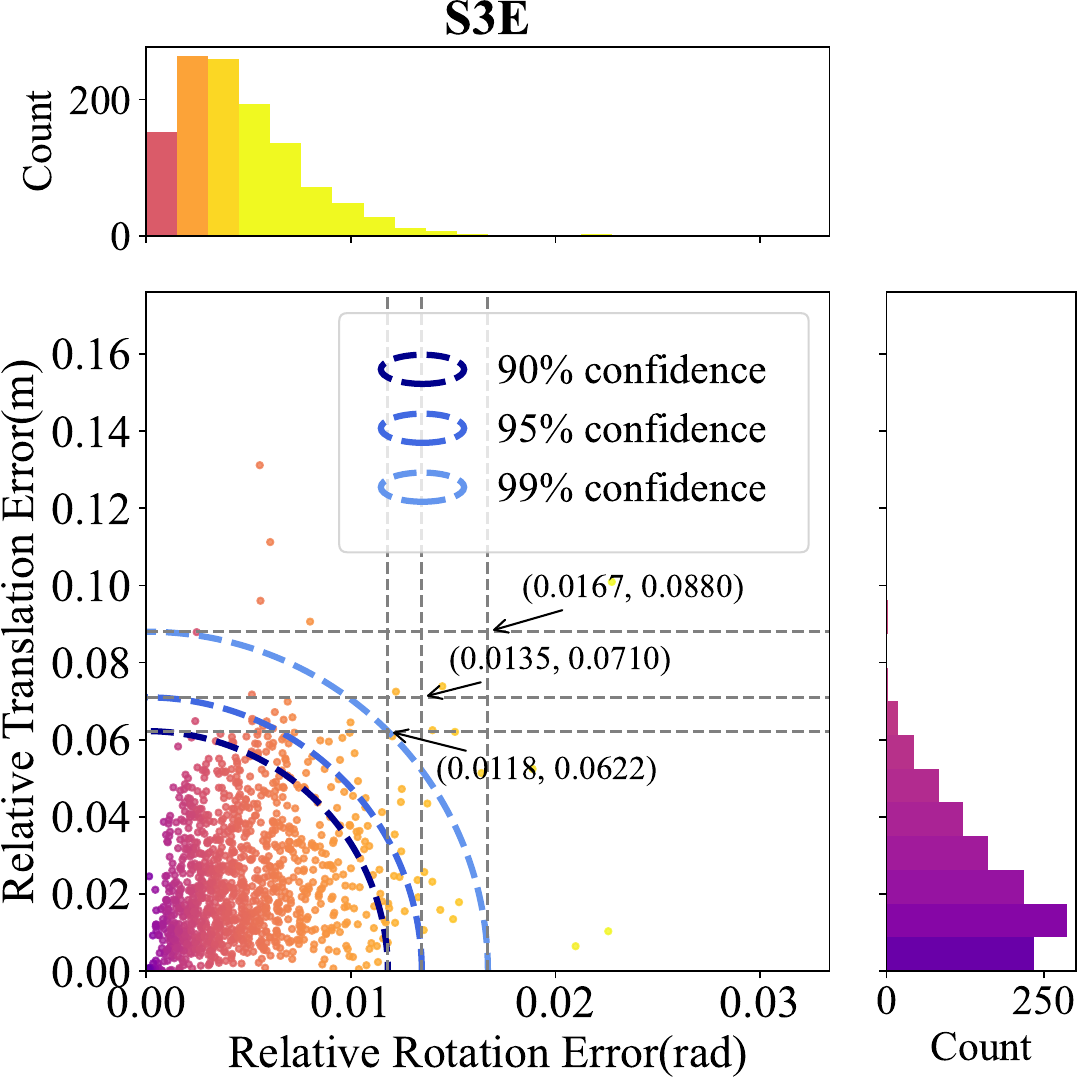}}
\caption{The relative rotation and translation error distributions, and quarter ellipse contours at 90\%, 95\%, and 99\% confidence levels for different methods.}
\label{fig_result_scatter1}
\end{figure*}
\begin{figure*}[!t]
\centering
{
    \includegraphics[width=0.185\linewidth]{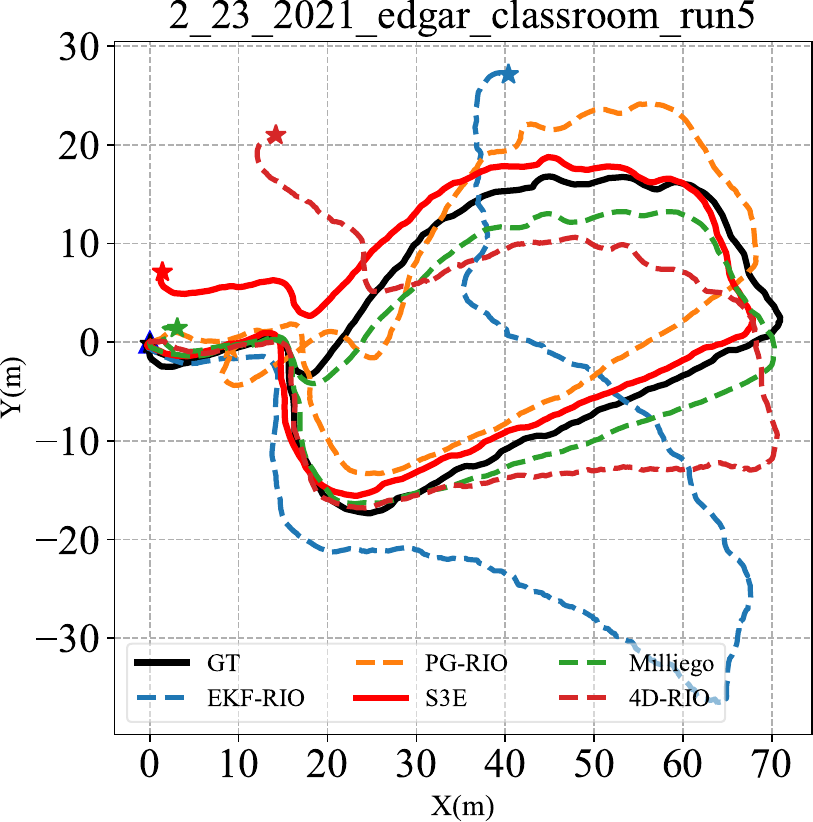}}
{
    \includegraphics[width=0.19\linewidth]{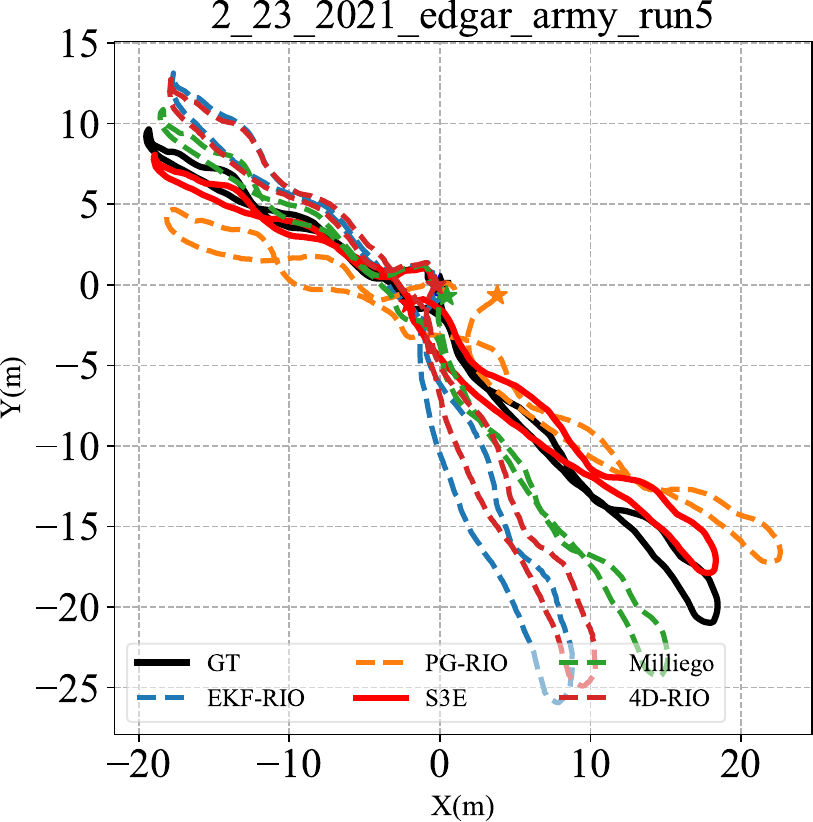}}
{ \includegraphics[width=0.185\linewidth]{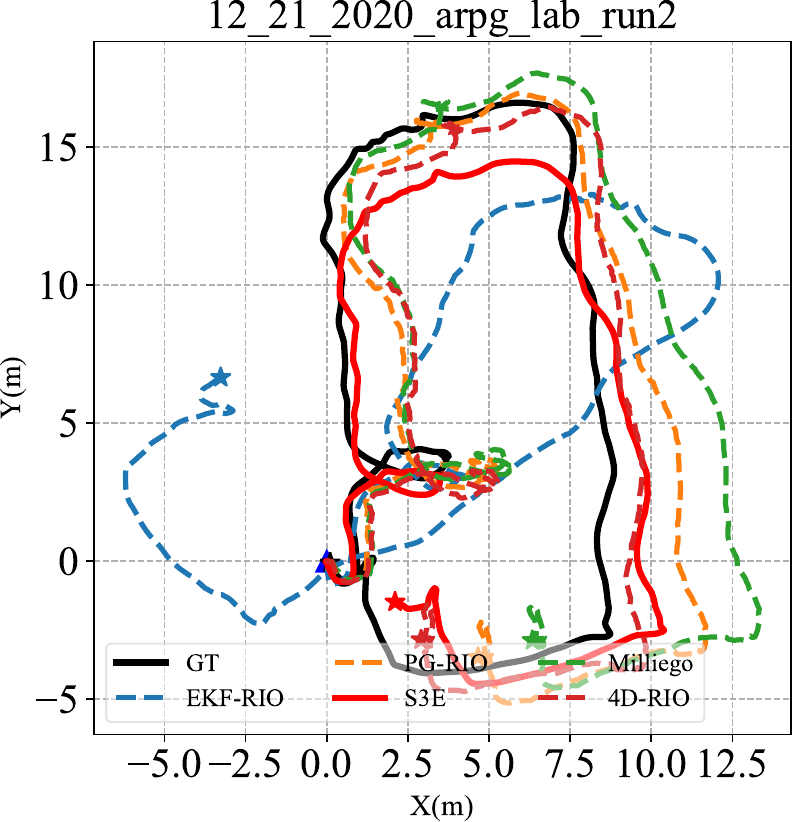}}
{
    \includegraphics[width=0.19\linewidth]{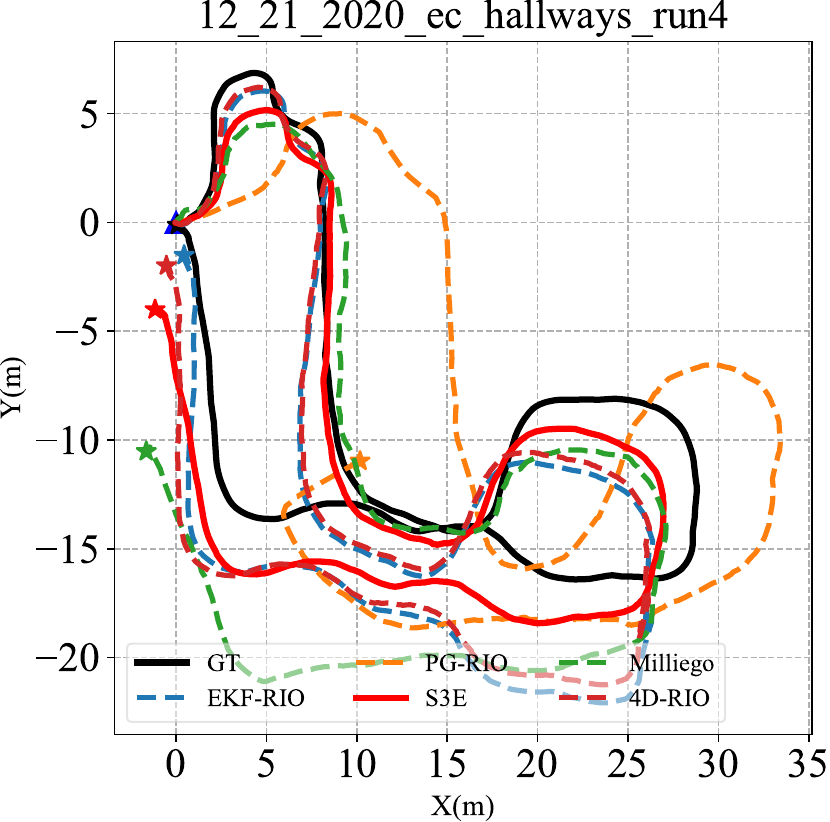}}
{
    \includegraphics[width=0.19\linewidth]{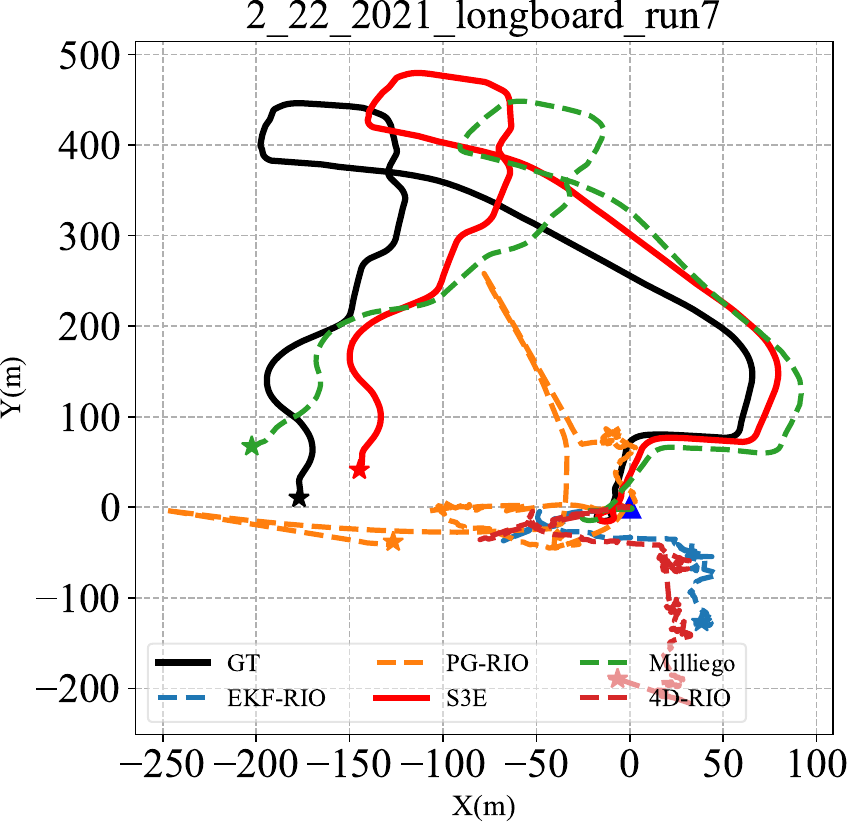}}
\caption{Pose estimation results are shown from different methods across various scenarios, where $\triangle$
indicates a unified initial point and $\star$ signifies the termination point of each respective algorithm.}
\label{fig_result_scatter2}
\end{figure*}

\begin{figure*}[!t]
\centering
{
    \includegraphics[width=0.19\linewidth]{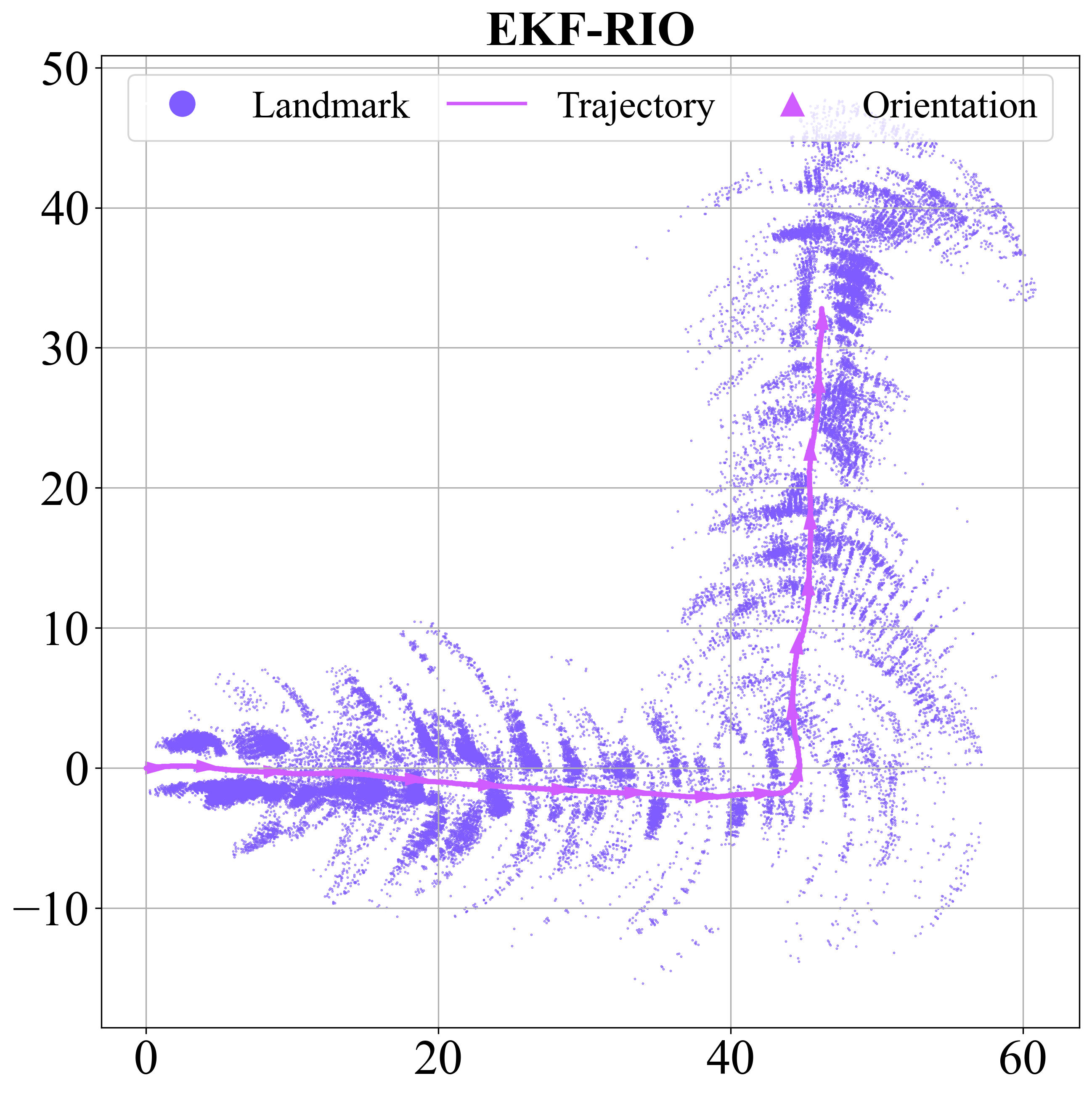}}
{
    \includegraphics[width=0.19\linewidth]{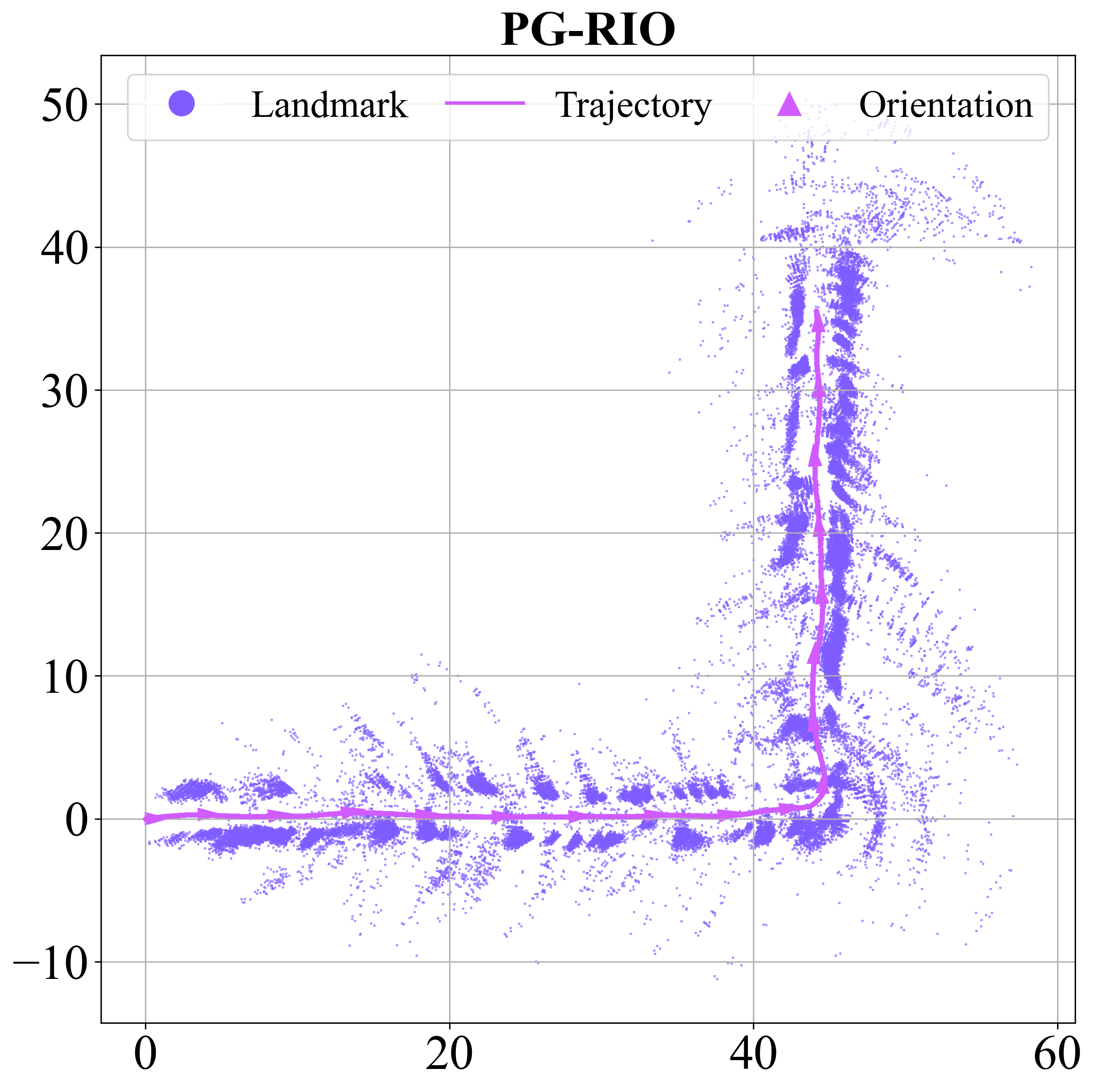}}
{
    \includegraphics[width=0.19\linewidth]{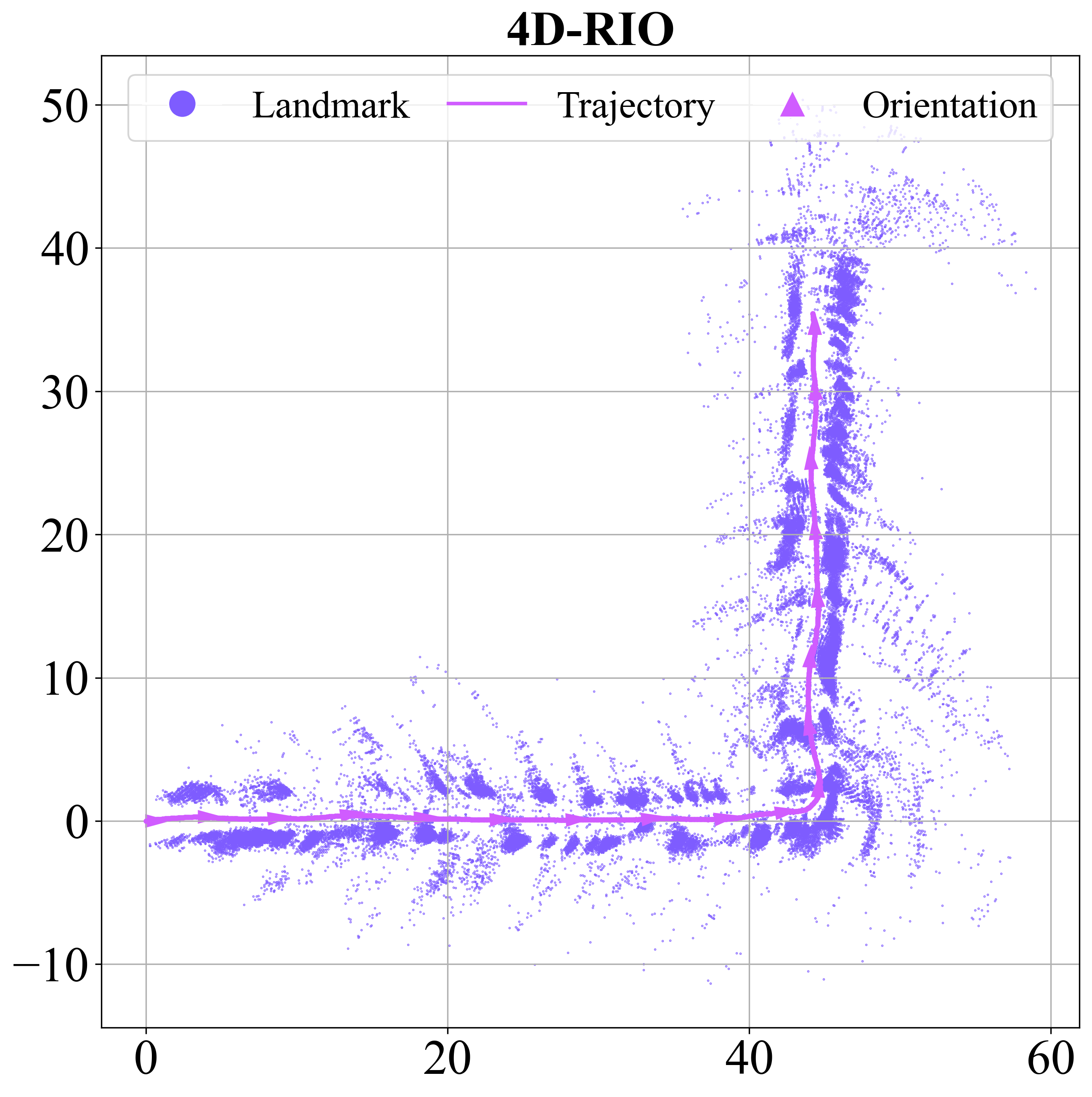}}
{
    \includegraphics[width=0.19\linewidth]{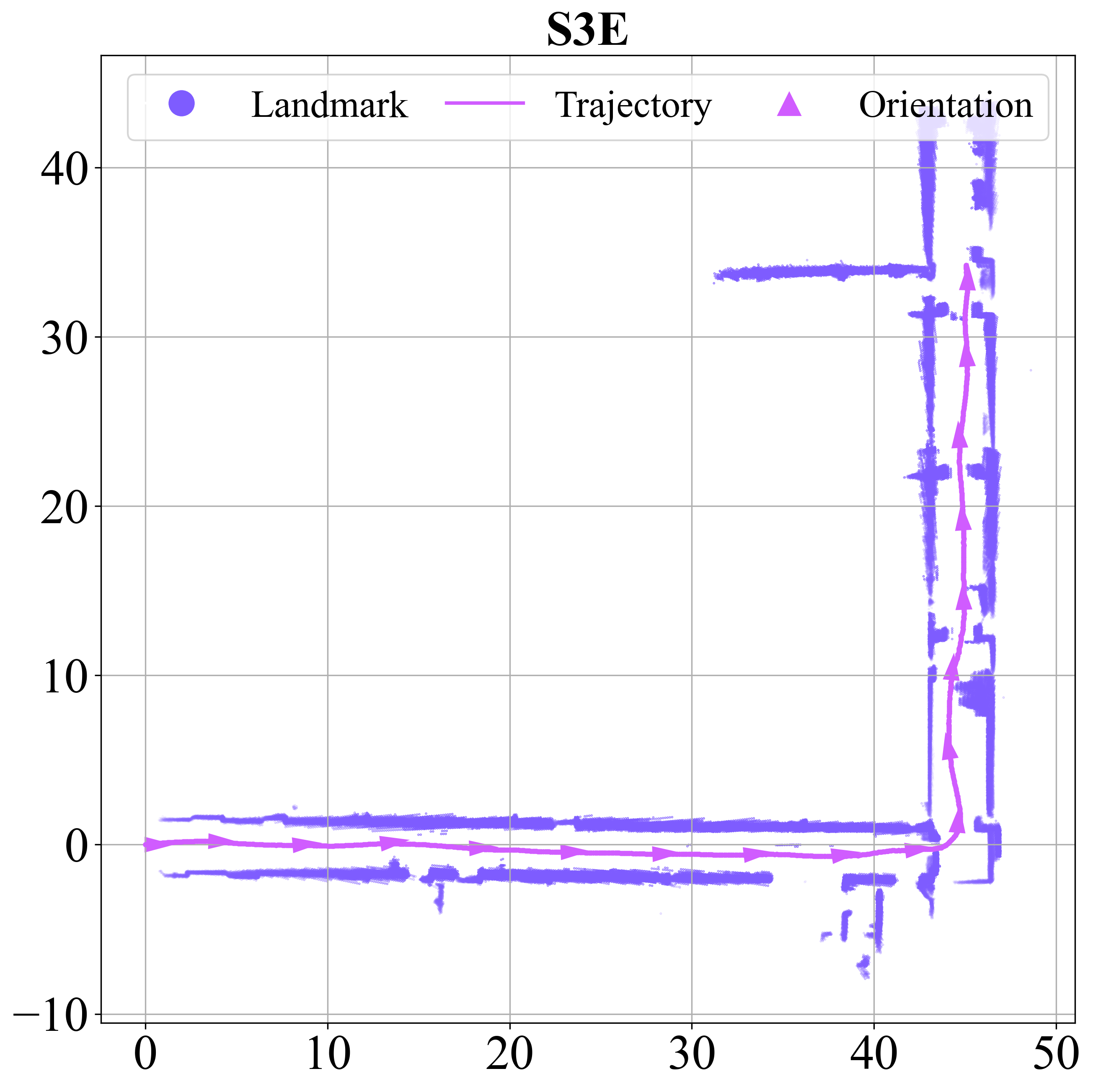}}
{
    \includegraphics[width=0.19\linewidth]{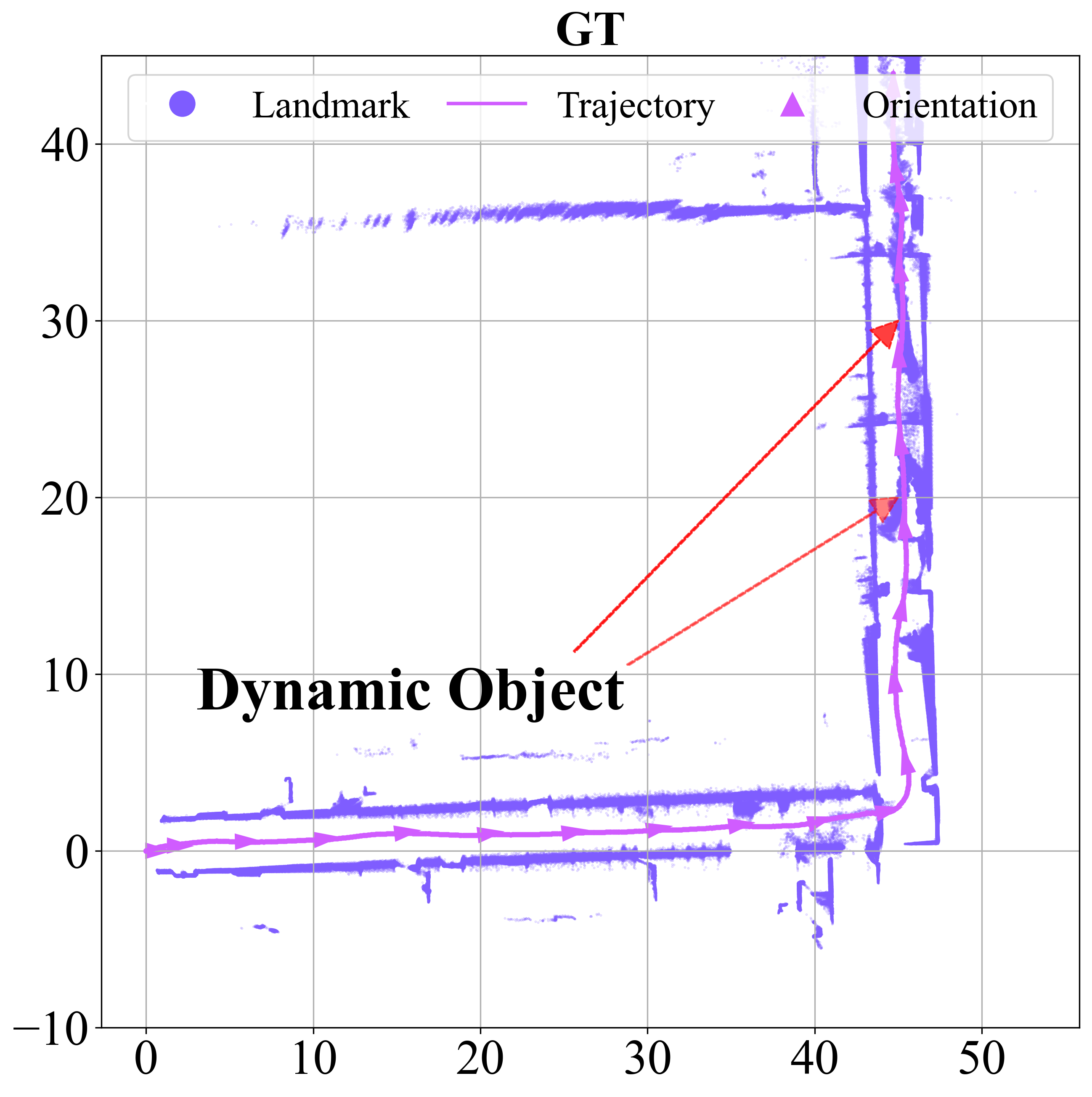}}
\caption{Pose and map visualizations from various methods and ground truth in self-collected dataset. All poses indicate
transformations relative to the initial coordinate system. Note that Milliego is unable to extract landmark point clouds.}
\label{fig_map}
\end{figure*}

\begin{table*}[h]
	\centering
	\fontsize{7}{8}\selectfont    
	\begin{tabular}{ccccccccccccccc}
		\toprule[0.5pt]
		\toprule
		\multirow{2}{*}{Methods}&\multicolumn{2}{c}{longboard} & \multicolumn{2}{c}{edgar\_classroom}& \multicolumn{2}{c}{outdoors} &\multicolumn{2}{c}{ec\_hallways}
		&\multicolumn{2}{c}{edgar\_army}&\multicolumn{2}{c}{arpg\_lab}&\multicolumn{2}{c}{aspen\_run}\cr
		\cmidrule(lr){2-3}
		\cmidrule(lr){4-5}
		\cmidrule(lr){6-7}
		\cmidrule(lr){8-9}
		\cmidrule(lr){10-11}
            \cmidrule(lr){12-13}
            \cmidrule(lr){14-15}
		& Trans.$\downarrow $ & Rot.$\downarrow $ & Trans.$\downarrow $ & Rot.$\downarrow $ & Trans.$\downarrow $ & Rot.$\downarrow $ & Trans.$\downarrow $ & Rot.$\downarrow $ & Trans.$\downarrow $ & Rot.$\downarrow $ & Trans.$\downarrow $ & Rot.$\downarrow $ & Trans.$\downarrow $ & Rot.$\downarrow $\cr
		\cmidrule(lr){1-15}
		EKF-RIO~\cite{doer2020ekf} & - & - & 5.32 & 6.08 & 4.65 & 5.61  & 5.68 & \textbf{2.16} & 6.19 & 5.38 &6.90&11.3 &9.73 &19.1 \cr
        PG-RIO~\cite{zhang20234dradarslam} & - & - & 2.61 & 2.99 & 7.67 & 14.7  & 5.58 & 5.26 & 3.94&3.99 &\textbf{2.31}&9.49 &2.74 & 15.8 \cr
        4D-RIO~\cite{xu2025incorporating} & - & - & 5.62 & 2.56 & 8.37 & 4.69 & 5.91 & 4.33 & 4.97 & 3.01 & 2.55 & 12.1 & 2.48 & 18.2 \cr
        Milliego~\cite{lu2020milliego} & 9.14 & 2.23  & \textbf{2.32} & \textbf{1.91} & \textbf{2.02} & 5.92 & 4.46 & 5.01 & \textbf{2.13}&2.45 &2.84&7.89 &3.17 &\textbf{11.0} \cr
       \rowcolor{gray50!20} S$^3$E & \textbf{5.69} & \textbf{2.07} & 2.45 & 2.21 & 2.16 & \textbf{3.83}  & \textbf{3.90} & 2.53 & 2.33& \textbf{2.11} &2.82&\textbf{4.28} &\textbf{2.34 }&11.3\cr
		\bottomrule
		\bottomrule[0.5pt]
	\end{tabular}\vspace{0cm}
    \caption{Comparison of rotation and translation error on the Coloradar Dataset. The rotation error~(abbreviated as Rot.) is measured in deg/100m and translation \textit{i.e.} Trans. in \%.}
	\label{tab:rpes}
\end{table*}

\begin{table*}[h]
	\centering
	\fontsize{7}{8}\selectfont    
	\begin{tabular}{ccccccccccccccc}
		\toprule[0.5pt]
		\toprule
		\multirow{2}{*}{Methods}&\multicolumn{2}{c}{indoor\_seq0} & \multicolumn{2}{c}{indoor\_seq1}& \multicolumn{2}{c}{indoor\_seq2} &\multicolumn{2}{c}{indoor\_seq3}
		&\multicolumn{2}{c}{outdoor\_seq0}&\multicolumn{2}{c}{outdoor\_seq1}&\multicolumn{2}{c}{outdoor\_seq2}\cr
		\cmidrule(lr){2-3}
		\cmidrule(lr){4-5}
		\cmidrule(lr){6-7}
		\cmidrule(lr){8-9}
		\cmidrule(lr){10-11}
            \cmidrule(lr){12-13}
            \cmidrule(lr){14-15}
		& Trans.$\downarrow $ & Rot.$\downarrow $ & Trans.$\downarrow $ & Rot.$\downarrow $ & Trans.$\downarrow $ & Rot.$\downarrow $ & Trans.$\downarrow $ & Rot.$\downarrow $ & Trans.$\downarrow $ & Rot.$\downarrow $ & Trans.$\downarrow $ & Rot.$\downarrow $ & Trans.$\downarrow $ & Rot.$\downarrow $\cr
		\cmidrule(lr){1-15}
		EKF-RIO~\cite{doer2020ekf} & 4.35 & 5.16 & 3.66 & 4.91 & 4.12 & 3.02  & 5.01 & 3.14 & - &- &-&- &- &- \cr
        PG-RIO~\cite{zhang20234dradarslam} & 3.68 & 4.10 & 5.27 & 3.65 & 4.32 & 3.71  & 4.25 & 2.96 &  - &- &-&- &- &- \cr
        4D-RIO~\cite{xu2025incorporating} & 5.65 & 3.23 & 5.34 & 3.17 & 4.16 & 2.79 & 4.62 & 2.77 &  - &- &-&- &- &- \cr
        Milliego~\cite{lu2020milliego} & 9.84 & 3.16  & 12.4 & 3.97 & 8.95 & 2.92 & 9.51 & 3.01 & 14.3 & 3.62 &16.9&3.76 &18.6 &3.55 \cr
       \rowcolor{gray50!20} S$^3$E & \textbf{2.69} & \textbf{2.13} & \textbf{2.97} & \textbf{2.41} &\textbf{ 2.35} & \textbf{1.98}  & \textbf{2.84} & \textbf{2.15} &\textbf{5.37}& \textbf{3.11} &\textbf{6.53}&\textbf{3.55} &\textbf{6.92} &\textbf{3.14}\cr
		\bottomrule
		\bottomrule[0.5pt]
	\end{tabular}\vspace{0cm}
	\caption{Evaluation on the fresh scenes, where the dataset is collected by our multi-modal sensing platform.}
	\label{tab:newscene}
\end{table*}

\subsection{Overall Performance}

We compare our method against model-based and learning-based localization approaches for the Radar-Inertial system in terms of translation and rotation error. For model-based methods, we utilize EKF-RIO~\cite{doer2020ekf}, PG-RIO~\cite{zhang20234dradarslam} and 4D-RIO~\cite{xu2025incorporating} as our baseline. As for learning-based methods, we adopt Milliego~\cite{lu2020milliego} as another baseline. In addition to absolute pose error, we portray the relative error distribution for a more intuitive evaluation of odometry.

Relative pose error distributions are analyzed to evaluate our odometry results against the baselines as illustrated in Fig. \ref{fig_result_scatter1}. The scatterplots show our method reduces relative rotation and translation errors by 11\% and 5.7\%, compared to Milliego, with a 90\% confidence level. This indicates more effective localization bias reduction and more reliable relative pose estimation.
For an overall trajectory as Fig. \ref{fig_result_scatter2} shows, our system
significantly contributes to a more accurate, less drifting trajectory. In addition, we evaluate the absolute pose error of different methods in various scenarios using the KITTI toolkit~\cite{Geiger2012CVPR}, as shown in Tab. \ref{tab:rpes}. S$^3$E still maintains superior performance in longboard scenarios where the CFAR detector generates fairly sparse point clouds. In contrast, EKF-RIO, PG-RIO and 4D-RIO fail due to inaccurate velocity estimation and poor point-wise matching. Among the scenes with valid evaluations, S$^3$E achieves the lowest average pose error, with a translation error of 3.10\% and a rotation error of 4.05 deg/100m, which demonstrates significant enhancement of 16.8\% in rotation and 22.2\% over supervised Milliego. This is attributed to more richly informative RAS rather than highly clustered points. Besides, S$^3$E in a self-supervised manner can be seamlessly adapted to new environments as shown in Tab.~\ref{tab:newscene}. Our approach consistently maintains robust and accurate localization performance.

\begin{figure}[htbp]
	\centering  
{
		\includegraphics[width=0.47\linewidth]{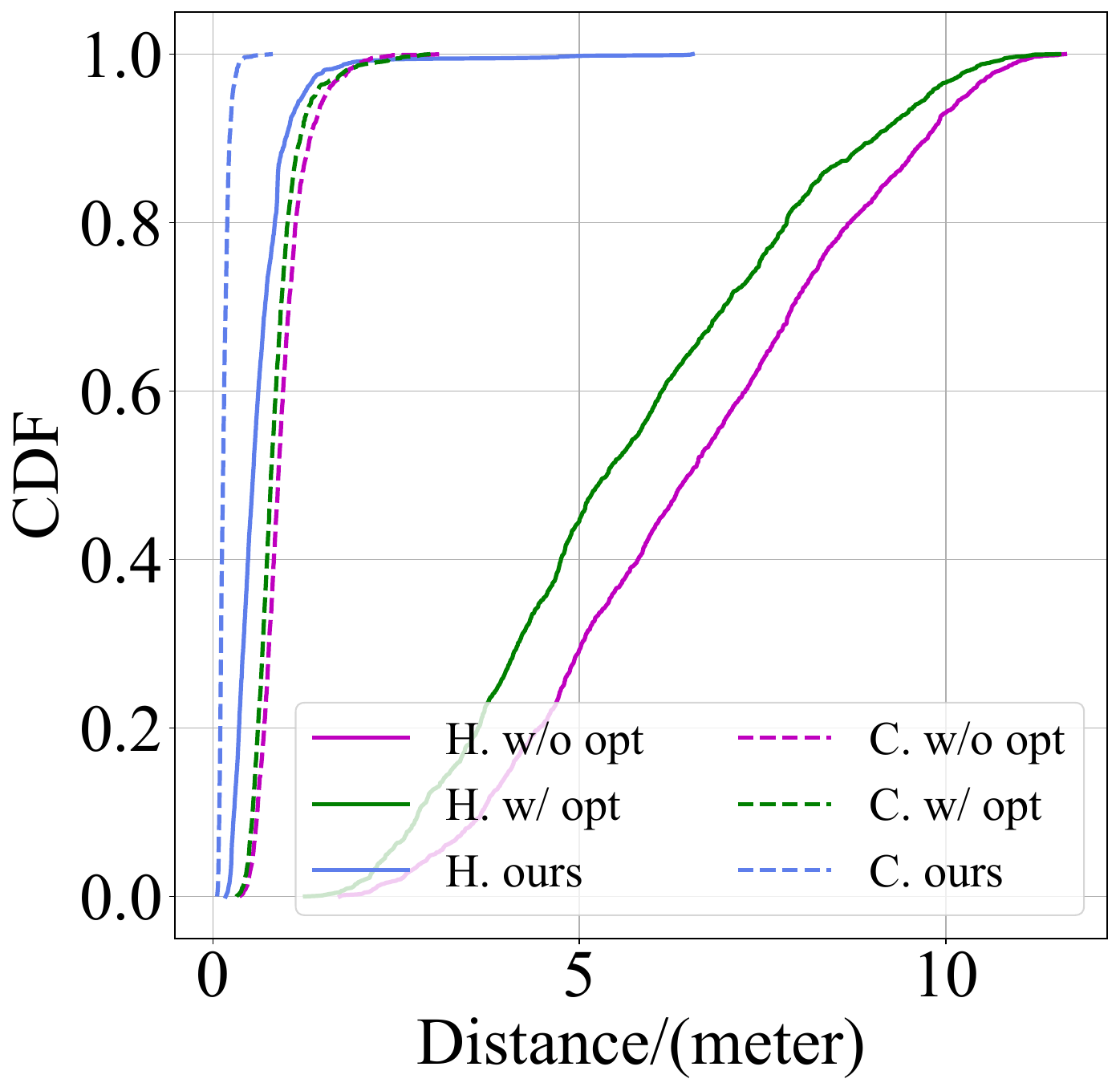}}
        \label{result_mapping_CDF}
{
		\label{result_mapping_RPCDL}
		\includegraphics[width=0.48\linewidth]{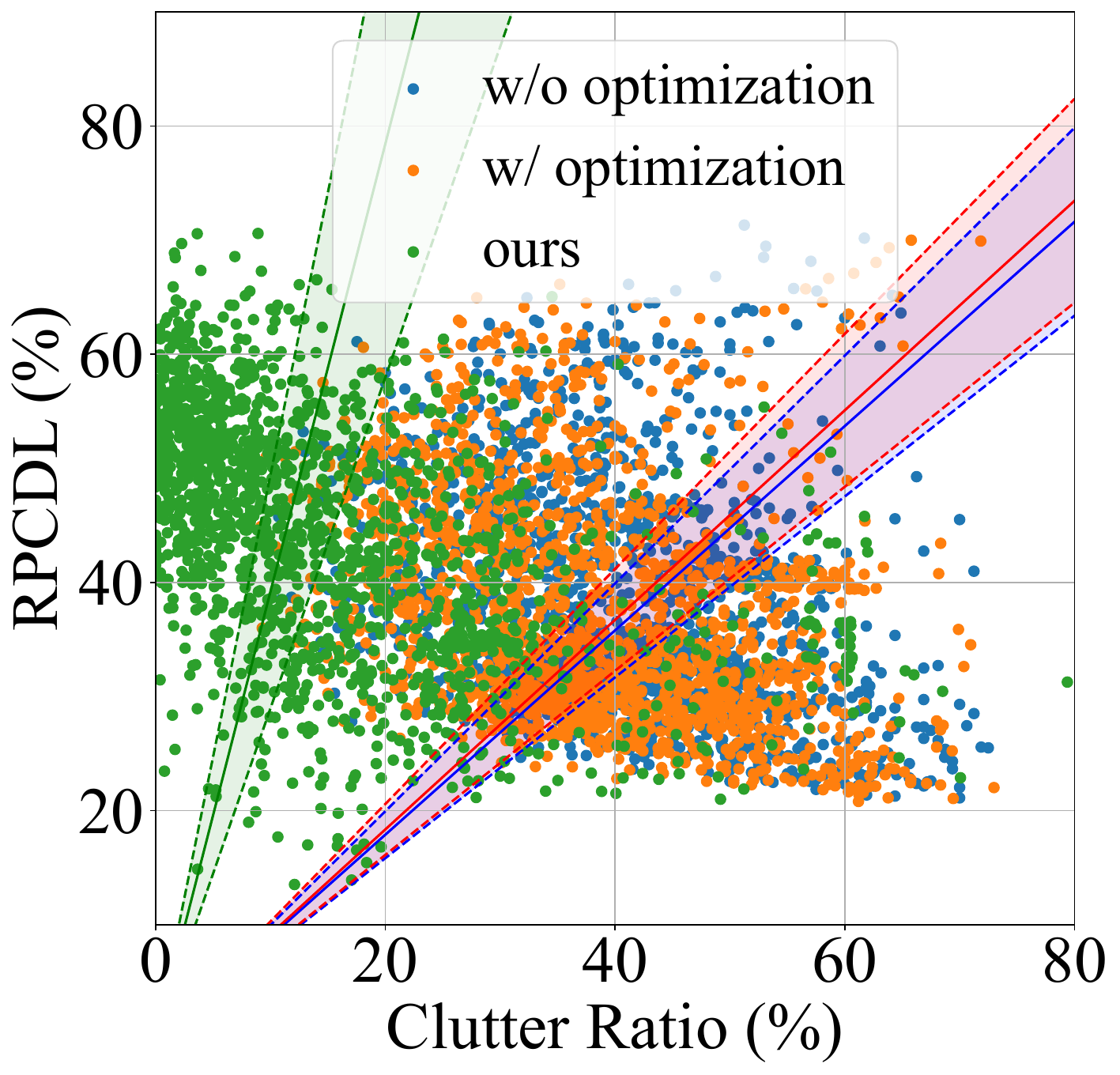}}
	\caption{Quality assessment of landmark extractors.}
	\label{result_mapping_eval}
\end{figure}

Furthermore, in contrast to Milliego, S$^3$E can additionally extract motion-consistent landmark features by leveraging the spatial structure information within RAS. Qualitatively, as shown in Fig.~\ref{fig_map}, our approach yields a more structured map with denser, more consistent landmark point clouds, and substantially fewer ghost points. Quantitatively, the performance of landmark extractors is evaluated using four metrics: Chamfer distance~\cite{guo2020deep}, Hausdorff distance~\cite{Hausdorff}, RPCDL~\cite{cheng2022novel} and Clutter Radio, as shown in Fig. \ref{result_mapping_eval}. The Chamfer and Hausdorff distances quantify the average distance between each point in two sets and its nearest neighbor in the opposite set. A higher RPCDL means the landmark point clouds are closer to the ground truth provided by LiDAR. Besides, a landmark point is defined as a clutter point if its nearest neighbor's distance from the ground truth exceeds 0.65m, and the Clutter Ratio indicates the fraction of landmark points identified as clutter. 

As shown in Fig.~\ref{result_mapping_eval}, we evaluate our landmark points against raw radar points used in EKF-RIO and optimized points by PG-RIO and 4D-RIO. The left shows that S$^3$E achieves the Chamfer and Hausdorff distances with
the median errors of 0.14m and 0.55m, respectively, which reduce to 10-17\% compared to baseline. The right illustrates the distribution of RPCDL and Clutter Ratio at different frames. Specifically, although PG-RIO filters and optimizes landmark points better than EKF-RIO, it still exhibits a higher Clutter Ratio and lower RPCDL compared to our method. Therefore, S$^3$E provides more geometry-consistent spatial landmark points with more accurate and robust perception than CFAR-based methods.

\subsection{Ablation Study}
To assess the individual contribution of the Rotation-based Cross Fusion module based on our insight in Fig. \ref{motivation}, we conduct an ablation experiment, presenting median errors for Chamfer Distance~(C.D.), Hausdorff Distance~(H.D.), and pose errors in translation and rotation, as shown in Tab. \ref{ablation}. The results indicate an 11-19\% improvement in state estimation and landmark extraction, demonstrating that Rotation-based Cross Fusion effectively preserves motion-consistent features and enhances the spatial structure between adjacent spectra.

\begin{table}[htbp!]
	\centering
\fontsize{8}{11}\selectfont
 \begin{threeparttable} 
\begin{tabular}{>{\centering\arraybackslash}m{0.5cm} 
		>{\centering\arraybackslash}m{0.5cm} 
		>{\centering\arraybackslash}m{0.8cm} 
		>{\centering\arraybackslash}m{0.8cm} 
		>{\centering\arraybackslash}m{1.3cm} 
		>{\centering\arraybackslash}m{1.7cm}}
	\toprule[0.5pt]
	\toprule
	S$^3$E&  RCF\tnote{$ \dagger $} & C.D(m)$\downarrow $ & H.D(m)$\downarrow $ & Trans.(\%)$\downarrow $& Rot.(deg/100m)$\downarrow $\cr
	\cmidrule(lr){1-6} 
	 $ \usym{2713} $& $ \usym{1F5F4} $ & 0.16 & 0.62 & 3.23& 2.57\cr
	 $ \usym{2713} $& $ \usym{2713} $  &  \textbf{0.14} & \textbf{0.55} & \textbf{2.84}& \textbf{2.15}\cr
	\bottomrule
	\bottomrule[0.5pt]
\end{tabular}\vspace{0cm}
 \begin{tablenotes} 
	\footnotesize     
	\item $ \dagger $ Rotation-based Cross Fusion in Fig. \ref{fig_2}.          
\end{tablenotes}
\end{threeparttable}
\caption{ABLATION STUDY RESULTS}
\label{ablation}
\end{table}

In a nutshell, our method S$^3$E thoroughly leverages geometric spatial consistency constraints while enhancing state estimation and landmark feature extraction in a complementary self-supervised manner, and two heterogeneous sensors synergize and complement each other, creating a harmonious interplay that enhances both. 
\section{Conclusion}
This paper introduces S$^3$E, a novel self-supervised state estimator and landmark extractor for Radar-Inertial Systems that leverages two key observations: subtle shifts in rotational components translate into linear shifts in the Range-Azimuth spectrum, and Doppler velocities can compensate for initial velocity, preventing cumulative IMU offsets. The experiments demonstrate that S$^3$E harnesses the complementary advantages of two heterogeneous sensors, achieving robust and accurate performance.

\section*{Acknowledgment}
This work was supported in part by National Natural Science Foundation of China with Grant 62471194.

{
    \small
    \bibliographystyle{ieeenat_fullname}
    \bibliography{main}
}

\end{document}